\pdfoutput=1 %
\documentclass{article} %
\PassOptionsToPackage{sort&compress}{natbib}
\usepackage{iclr2027_conference,times}
\setcitestyle{numbers,square,citesep={,}}

\usepackage{amsmath,amsfonts,bm}

\def\eqref#1{equation~\ref{#1}}

\def\1{\bm{1}}

\DeclareMathAlphabet{\mathsfit}{\encodingdefault}{\sfdefault}{m}{sl}
\SetMathAlphabet{\mathsfit}{bold}{\encodingdefault}{\sfdefault}{bx}{n}

\definecolor{GeoReaderBurntOrange}{RGB}{255,142,38}
\definecolor{GeoReaderEmerald}{RGB}{1,177,160}
\definecolor{TableGroupFill}{RGB}{238,238,238}
\definecolor{TableSecondFill}{RGB}{228,228,228}
\definecolor{TableBestFill}{RGB}{255,240,222}
\newcommand{\bestkey}{\begingroup\setlength{\fboxsep}{1pt}\colorbox{TableBestFill}{\textbf{bold}}\endgroup}
\newcommand{\secondbestkey}{\begingroup\setlength{\fboxsep}{1pt}\colorbox{TableSecondFill}{gray}\endgroup}
\newcommand{\tablegrouprule}{\specialrule{\lightrulewidth}{0pt}{0pt}}
\newcommand{\tablegroupgap}{\addlinespace[2pt]}

\makeatletter
\newsavebox{\tablepadbox}
\newcommand{\tablepadabove}[1]{\sbox{\tablepadbox}{\copy\@arstrutbox#1}\rule{0pt}{\dimexpr\ht\tablepadbox+\belowrulesep\relax}\usebox{\tablepadbox}}
\newcommand{\tablepadbelow}{\rule[-\dimexpr\dp\@arstrutbox+\aboverulesep\relax]{0pt}{\dimexpr\dp\@arstrutbox+\aboverulesep\relax}}
\makeatother
\newcommand{\filledtoprule}{\specialrule{\heavyrulewidth}{\abovetopsep}{0pt}}
\newcommand{\filledmidrule}{\specialrule{\lightrulewidth}{0pt}{0pt}}
\newcommand{\filledbottomrule}{\specialrule{\heavyrulewidth}{0pt}{\belowbottomsep}}

\definecolor{GeoReaderWordmarkOrange}{rgb}{0.953,0.298,0.0314}
\definecolor{GeoReaderWordmarkGreen}{rgb}{0,0.553,0.439}
\newcommand{\georeaderplain}{GeoReader}
\newif\ifgeoreaderbranding
\georeaderbrandingtrue
\newcommand{\georeadertitle}{%
  \texorpdfstring{%
    \ifgeoreaderbranding
      \textcolor{GeoReaderWordmarkOrange}{Geo}%
      \textcolor{GeoReaderWordmarkGreen}{Reader}%
    \else
      \georeaderplain%
    \fi
  }{\georeaderplain}%
}
\newcommand{\georeader}{%
  \texorpdfstring{%
    \textbf{%
      \ifgeoreaderbranding
        \textcolor{GeoReaderWordmarkOrange}{Geo}%
        \textcolor{GeoReaderWordmarkGreen}{Reader}%
      \else
        \georeaderplain%
      \fi
    }%
  }{\georeaderplain}%
}
\newcommand{\georeaderfig}{%
  \ifgeoreaderbranding
    \textcolor{GeoReaderWordmarkOrange}{Geo}%
    \textcolor{GeoReaderWordmarkGreen}{Reader}%
  \else
    \georeaderplain%
  \fi
}
\newcommand{\ours}{\georeader}
\newsavebox{\roundedimagebox}
\newcommand{\roundedimage}[1]{%
  \sbox{\roundedimagebox}{#1}%
  \begin{tikzpicture}
    \clip[rounded corners=2pt] (0,0) rectangle (\wd\roundedimagebox,\ht\roundedimagebox);
    \node[anchor=south west,inner sep=0pt,outer sep=0pt] at (0,0) {\usebox{\roundedimagebox}};
    \draw[draw=black!18,line width=0.8pt,rounded corners=2pt]
      (0.4pt,0.4pt) rectangle (\dimexpr\wd\roundedimagebox-0.4pt\relax,\dimexpr\ht\roundedimagebox-0.4pt\relax);
  \end{tikzpicture}}

\newcommand{\figkeyword}[1]{{\setlength{\fboxsep}{0.7pt}\colorbox[HTML]{FFF0A8}{\strut #1}}}

\newcommand{\correctanswer}[1]{\textcolor{black!70}{\tikz[baseline=-0.4pt]\draw[sharp corners,line width=1.3pt,line cap=round,line join=round] (0pt,2pt)--(2pt,0pt)--(6pt,5pt);}\,\textcolor{black!70}{#1}}

\usepackage{graphicx}
\usepackage{tikz}
\usepackage[T1,OT1]{fontenc}
\usepackage{booktabs}
\usepackage{colortbl}
\usepackage{tabularx} %
\usepackage{multirow}
\usepackage{placeins}
\usepackage{afterpage}
\newcommand{\headingfont}{\fontencoding{T1}\fontfamily{lato-TLF}\bfseries\selectfont}
\makeatletter
\renewcommand{\section}{\@startsection{section}{1}{\z@}{-2.0ex plus -0.5ex minus -.2ex}{1.5ex plus 0.3ex minus 0.2ex}{\headingfont\large\raggedright\color{GeoReaderWordmarkGreen}}}
\renewcommand{\subsection}{\@startsection{subsection}{2}{\z@}{-1.8ex plus -0.5ex minus -.2ex}{0.8ex plus .2ex}{\headingfont\normalsize\raggedright\color{GeoReaderWordmarkGreen}}}
\renewcommand{\subsubsection}{\@startsection{subsubsection}{3}{\z@}{-1.5ex plus -0.5ex minus -.2ex}{0.5ex plus .2ex}{\headingfont\normalsize\raggedright\color{GeoReaderWordmarkGreen}}}
\renewcommand{\paragraph}{\@startsection{paragraph}{4}{\z@}{1.5ex plus 0.5ex minus .2ex}{-1em}{\headingfont\normalsize\color{GeoReaderWordmarkGreen}}}
\makeatother
\definecolor{TableGroupFill}{RGB}{238,244,239}
\definecolor{TableGroupText}{RGB}{20,94,76}
\newcommand{\tablegroup}[1]{\rlap{\setlength{\fboxsep}{0pt}\colorbox{TableGroupFill}{\makebox[\linewidth][l]{\rule[-0.9ex]{0pt}{3.1ex}\hspace{4pt}{\color{GeoReaderWordmarkGreen}\rule[0.1ex]{0.85ex}{0.85ex}}\hspace{0.45em}{\headingfont\itshape\fontsize{8}{9.5}\selectfont\color{TableGroupText}#1}}}}}
\usepackage{caption}
\DeclareCaptionFont{heading}{\headingfont}
\usepackage{hyperref}
\hypersetup{colorlinks=true,linktoc=all,linkcolor=GeoReaderWordmarkGreen,citecolor=GeoReaderWordmarkGreen,urlcolor=GeoReaderWordmarkGreen,filecolor=GeoReaderWordmarkGreen,hypertexnames=false,pdftitle={GeoReader: Learning Latent Geometry for Spatial Reasoning in Vision-Language Models},pdfauthor={Rong Li, Shijie Li, Jiangyong Huang, Lingdong Kong, Jiaming Zhou, Dongyue Lu, Zeying Gong, Baoxiong Jia, Junwei Liang, Siyuan Huang}}
\usepackage[nameinlink]{cleveref}
\crefname{section}{Sec.}{Secs.}
\crefname{subsection}{Sec.}{Secs.}
\crefname{subsubsection}{Sec.}{Secs.}
\crefname{appendix}{Sec.}{Secs.}
\crefname{subappendix}{Sec.}{Secs.}
\crefname{subsubappendix}{Sec.}{Secs.}
\crefname{table}{Tab.}{Tabs.}
\crefname{figure}{Fig.}{Figs.}
\crefname{equation}{Eq.}{Eqs.}
\Crefname{section}{Sec.}{Secs.}
\Crefname{subsection}{Sec.}{Secs.}
\Crefname{subsubsection}{Sec.}{Secs.}
\Crefname{appendix}{Sec.}{Secs.}
\Crefname{subappendix}{Sec.}{Secs.}
\Crefname{subsubappendix}{Sec.}{Secs.}
\Crefname{table}{Tab.}{Tabs.}
\Crefname{figure}{Fig.}{Figs.}
\Crefname{equation}{Eq.}{Eqs.}
\creflabelformat{equation}{#2#1#3}

\usepackage{url}

\newcommand{\linkfont}{\fontencoding{T1}\fontfamily{lato-TLF}\selectfont}
\usepackage{fontawesome5} %

\title{\upshape\bfseries\georeadertitle:\hspace{0.2em}Learning Latent Geometry for\\Spatial Reasoning in Vision-Language Models}

\iclrfinalcopy %

\begin{document}
\addtocontents{toc}{\protect\setcounter{tocdepth}{-10}}

\definecolor{TitleCardFill}{rgb}{0.945,0.957,0.969}
\newcommand{\affiliationlogos}{\includegraphics[height=22pt]{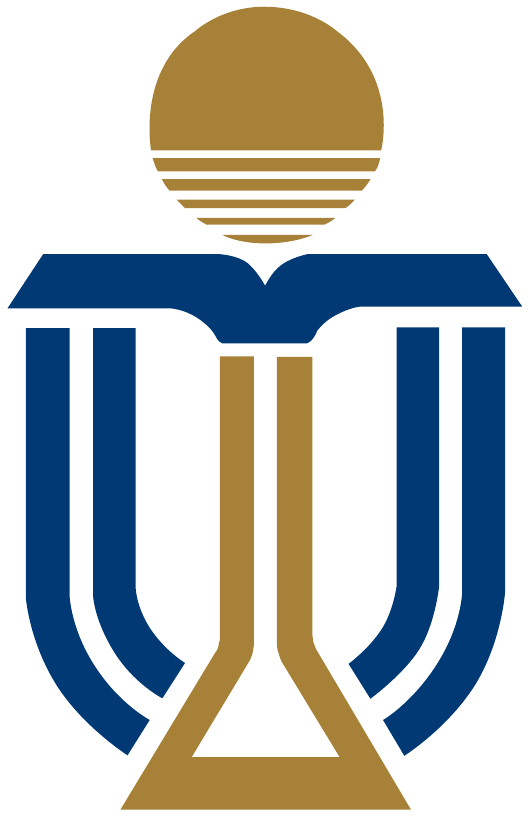}\hspace{7pt}\includegraphics[height=22pt]{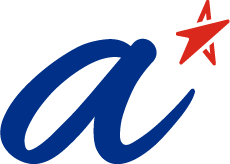}\hspace{7pt}\includegraphics[height=22pt]{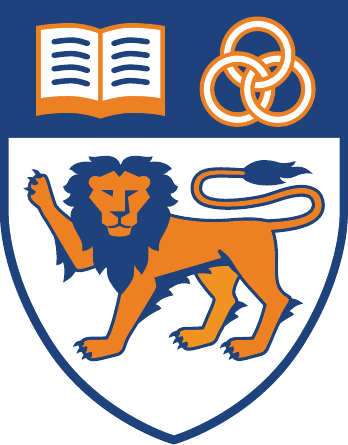}\hspace{7pt}\includegraphics[height=22pt]{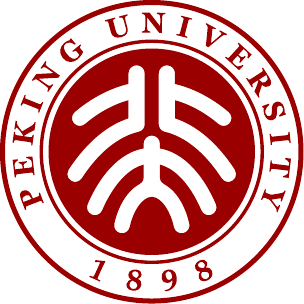}\hspace{7pt}\includegraphics[height=22pt]{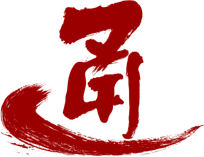}}
\lhead{}
\renewcommand{\headrulewidth}{0pt}
\fancypagestyle{arxivfirstpage}{%
  \fancyhf{}%
  \renewcommand{\headrulewidth}{0pt}%
  \fancyfoot[C]{\thepage}%
}
\thispagestyle{arxivfirstpage}
\renewenvironment{abstract}{}{\par}
\noindent\hspace*{-14pt}%
\begin{tikzpicture}
\node[fill=TitleCardFill,rounded corners=14pt,inner xsep=14pt,inner ysep=16pt]{%
\begin{minipage}{\linewidth}
{\LARGE\headingfont\georeadertitle}\hfill\affiliationlogos\par\vspace{8pt}
{\large\headingfont Learning Latent Geometry for Spatial Reasoning in Vision-Language Models\par}\vspace{12pt}
\textbf{Rong Li$^{1,6*}$\quad Shijie Li$^{2*}$\quad Jiangyong Huang$^{3,6}$\quad Lingdong Kong$^{4}$\quad Jiaming Zhou$^{1}$}\\
\textbf{Dongyue Lu$^{4}$\quad Zeying Gong$^{1}$\quad Baoxiong Jia$^{6\dagger}$\textsuperscript{\faEnvelope[regular]}\quad Junwei Liang$^{1,5}$\textsuperscript{\faEnvelope[regular]}\quad Siyuan Huang$^{6}$\textsuperscript{\faEnvelope[regular]}}\\[4pt]
{\normalfont\small $^{1}$The Hong Kong University of Science and Technology (Guangzhou)}\\
{\normalfont\small $^{2}$Agency for Science, A*STAR\quad $^{3}$Peking University}\\
{\normalfont\small $^{4}$National University of Singapore\quad $^{5}$The Hong Kong University of Science and Technology}\\
{\normalfont\small $^{6}$Beijing Institute for General Artificial Intelligence}\par\vspace{12pt}
{\large\headingfont\color{GeoReaderWordmarkGreen} Abstract\par}\vspace{4pt}
\begin{abstract}
Vision-language models (VLMs) struggle with metric and multi-view relationships
that remain implicit in RGB observations. Geometry foundation models recover this structure from RGB, but existing methods either fuse the estimated features with
visual tokens and learn to use them from question answering alone, or align VLM latents with geometry features. Capturing geometry, however, does not ensure its use: zeroing the aligned latents of a released feature-alignment model leaves its accuracy unchanged. We introduce \georeader, a framework that learns geometric representations
for the VLM that reads them. First, we train lightweight Geometric Readers to produce continuous input tokens, supervising them with geometry descriptions decoded by a frozen VLM, so that the Readers express geometry in a form the target model can already read. We then freeze the Readers to preserve
this interface while fine-tuning the VLM on spatial QA, with geometry injection
conditioned on the question. Geometry descriptions serve only as training
targets: at inference, the VLM answers directly from the continuous geometric tokens without
generating a geometry chain of thought. %
With geometry predicted from RGB images, \georeader{} reaches $94.3$ on
MindCube-Tiny with Qwen3-VL-4B and, with a multi-source QA mixture, an average
of $72.5$ across six additional spatial benchmarks with Qwen3.5-9B. These results
suggest that a useful geometric representation is one the receiving model can
use, not merely one from which geometry can be reconstructed.
\end{abstract}

\vspace{10pt}
{\small\faGlobe\hspace{5pt}{\headingfont Project Page:}\hspace{5pt}\textcolor{GeoReaderWordmarkGreen}{\url{https://iris0329.github.io/georeader-homepage/}}\par}\vspace{3pt}
{\small\faGithub\hspace{5pt}{\headingfont GitHub Repo:}\hspace{5pt}\textcolor{GeoReaderWordmarkGreen}{\url{https://github.com/iris0329/GeoReader}}\par}
\end{minipage}};
\end{tikzpicture}\hspace*{-14pt}\par
\vspace{14pt}
{\renewcommand{\thefootnote}{}\footnotetext{$^{*}$Equal contribution.\quad \textsuperscript{\faEnvelope[regular]}Corresponding authors.\quad $^{\dagger}$Project lead.}}

\begin{figure}[t]
  \centering
  \includegraphics[width=\linewidth]{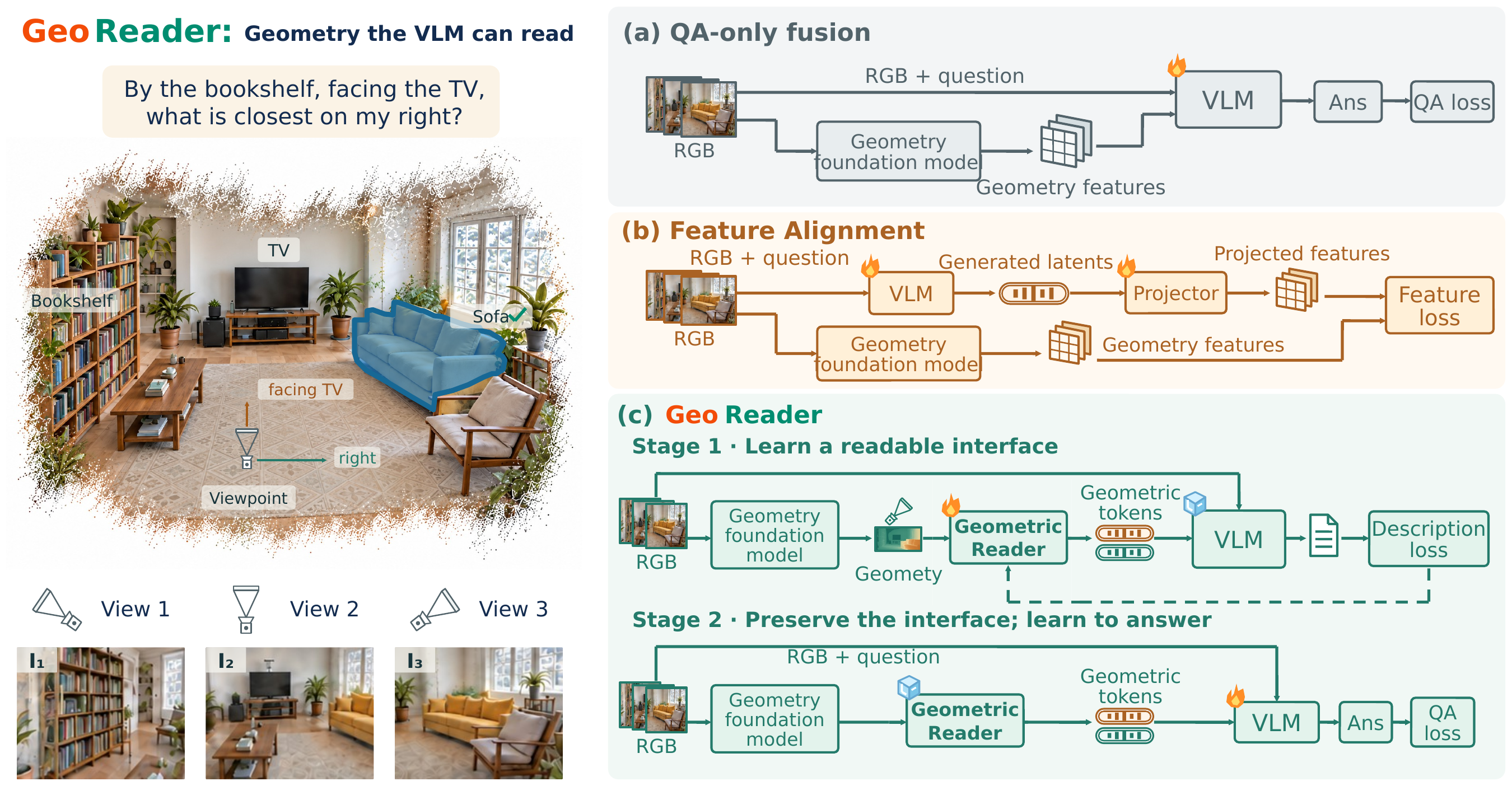}
  \caption{\georeader{} learns geometry in a form the receiving VLM can read.
  (a) QA-only fusion learns to use geometry through answer supervision.
  (b) Feature alignment matches VLM-generated representations to external
  geometry features. (c) \georeader{} first trains Geometric Readers through
  description prediction by a frozen VLM, then freezes the Readers and trains
  the VLM for QA. All three routes estimate geometry
  from RGB, which also directly conditions each VLM. Flames and ice cubes
  mark trainable and frozen modules. The pictured $I_1,I_2,I_3$ are input
  views; the scene and answer are illustrative. \cref{fig:method-overview}
  details the architecture.}
  \label{fig:teaser}
\end{figure}

\section{Introduction}
\label{sec:introduction}

Vision-language models (VLMs) have demonstrated increasingly strong
capabilities in general multimodal understanding, yet continue to struggle with
spatial reasoning, a capability essential to embodied intelligence in the 3D
world~\citep{li2025seeground,hu2026spark}. A central challenge is inferring and reasoning about 3D structure from
2D observations~\citep{li2025gamonossc}, as required by spatial questions involving camera motion,
viewpoint changes, and object distances. Recent advances in geometry foundation
models~\citep{wang2025vggt,lin2025da3} offer a promising opportunity to address
this limitation by providing geometry information. How to harness
this information for VLM spatial reasoning has
therefore become an important question.

Existing approaches to integrating geometry foundation models with VLMs have
improved spatial reasoning and can be broadly categorized into two streams. The
first introduces geometry as an additional input
modality~\citep{fan2025vlm3r,wu2025spatialmllm,zhao2025spacemind}, fusing features from geometry foundation models with the VLM's original visual features (\cref{fig:teaser}a). This provides a direct way to incorporate geometry into existing VLMs,
and the VLM learns to use these features from downstream supervision.
However, this supervision alone may not teach the VLM to
interpret geometry features. The second uses geometry
as supervision, typically by aligning VLM latent representations with features from geometry foundation models~\citep{chen2026think3d} (\cref{fig:teaser}b). This encourages the
latent space to capture geometry information while preserving the VLM's input
and output interfaces. Nevertheless, capturing geometry does not ensure its use: zeroing the latents of the released first-stage 3DThinker~\citep{chen2026think3d} checkpoint leaves MindCube-Tiny~\citep{yin2025mindcube} accuracy essentially unchanged, and answer tokens barely attend to them (\cref{tab:token-use-ablation}; \cref{fig:token-attention}).
Related analyses of latent visual reasoning methods such as
LVR~\citep{li2026lvr} and Monet~\citep{wang2026monet} also highlight a
disconnect between intermediate representations and answer
generation~\citep{viveiros2026holding,guo2026beyond,li2026imagination,zhu2026silence,park2026cvrr}.
Despite their differences, neither approach explicitly bridges the gap between
geometry representations and language-grounded spatial relations for VLMs.
Consequently, the role of geometry information in the model's reasoning remains
uncertain. We bridge this gap by converting geometry into a form that VLMs can both interpret and ground in language.

We introduce \georeader, based on the core idea that \emph{geometry should first be converted into a form that VLMs can interpret as spatial relationships in language, before VLMs learn to use it for downstream tasks}. To this end, a two-stage pipeline separates adapting geometry to the VLM from applying it to spatial QA (\cref{fig:teaser}c). In Stage~1,
we train lightweight Geometric Readers to convert per-view geometric signals
into continuous geometric tokens. These tokens and the corresponding images are given to a frozen VLM that predicts geometry descriptions. With
the VLM frozen, this supervision trains the Readers to express geometry
information in a form that the model can interpret through its existing
language capabilities. In Stage~2, we transfer the pretrained Readers to spatial QA, freezing the Readers to preserve their learned mapping while fine-tuning the VLM to use their outputs. Since
questions differ in how much they rely on each geometric signal, a geometry injector conditions on the images and question to set each signal's strength. Geometry descriptions are used only for Stage~1
supervision; at inference, the VLM answers directly using the geometric tokens.

We evaluate \georeader{} across seven spatial reasoning benchmarks and three
VLM families. With geometry predicted from RGB images, \georeader{} achieves
$94.3\%$ accuracy on MindCube-Tiny with Qwen3-VL-4B and, with multi-source QA
training, an average accuracy of $72.5\%$ across six additional benchmarks with
Qwen3.5-9B (\cref{sec:spatial-performance}). Comparisons with vanilla
supervised fine-tuning (SFT) under matched QA data and training settings show
gains concentrated on questions requiring metric 3D understanding. %
Ablations
further show the benefits of readability pretraining, freezing the
Readers during downstream adaptation, and the geometry injector (\cref{sec:readability-evaluation,sec:utilization-evaluation}).
Geometry interventions show that the model uses the supplied geometry: in a controlled MindCube-Tiny experiment with annotation-derived
camera geometry, removing geometry content reduces accuracy from $89.2\%$ to
$77.2\%$, while shuffling poses across views reduces it to $63.1\%$.

Our contributions are summarized as follows:
\begingroup
\setlength{\leftmargini}{1.3em}
\setlength{\topsep}{3pt}
\setlength{\partopsep}{0pt}
\begin{itemize}
    \setlength{\itemsep}{2pt}
    \setlength{\parsep}{0pt}
    \setlength{\parskip}{0pt}
    \item We propose explicitly connecting geometry information to spatial
    relationships expressed in language, separating adapting geometry to the VLM from applying it to spatial reasoning.
    \item We introduce \georeader{}, a two-stage framework that trains lightweight
    Geometric Readers through language supervision from a frozen VLM, then
    freezes the Readers and adapts the VLM for spatial QA with
    question-conditioned geometry injection (\cref{sec:method}).
    \item We demonstrate the effectiveness of \georeader{} across seven spatial
    reasoning benchmarks and three VLM families, with ablations and geometry
    interventions supporting both the framework's design and the model's use of
    geometry information (\crefrange{sec:spatial-performance}{sec:interface-tradeoffs}).
\end{itemize}
\endgroup

\section{Related Work}
\label{sec:related-work}

\paragraph{Spatial reasoning in vision-language models.}
SpatialVLM~\citep{chen2024spatialvlm}, SpaRE~\citep{spare2025} and
SPAR-7M~\citep{zhang2025spar} build synthetic or 3D-grounded spatial QA data.
SpatialLadder~\citep{li2025spatialladder} and HiSpatial~\citep{liang2026hispatial}
organize training from perception toward reasoning, while
SenseNova-SI~\citep{sensenova2025} studies scaling spatial data.
SpatialQA~\citep{spatialbot2024} and ViewSpatial~\citep{li2025viewspatial}
target depth and viewpoint relations; N3D-VLM~\citep{n3d2025} links 3D object localization
to reasoning.
These works establish the value of spatial supervision; we study how to present estimated geometry to the answering VLM.

\paragraph{Geometry-augmented VLMs.}
DUSt3R~\citep{wang2023dust3r} and MASt3R~\citep{leroy2024mast3r} estimate point maps and correspondences;
CUT3R~\citep{wang2025cut3r} maintains a recurrent scene state.
VGGT~\citep{wang2025vggt} predicts camera and dense geometry, and
Depth Anything V2~\citep{yang2024depthv2} and Depth Anything 3~\citep{lin2025da3}
predict depth and any-view geometry.
Geometry-augmented VLMs take such 3D structure as input: 3D-LLM~\citep{hong2023llm3d}
and LEO~\citep{huang2023leo} ground language in 3D scenes.
LLaVA-3D~\citep{zhu2024llava3d}, Video-3D LLM~\citep{zheng2024video3d} and
SR-3D~\citep{sr3d2025} add spatial positions or regions to visual inputs.
SpatialRGPT~\citep{cheng2024spatialrgpt}, SD-VLM~\citep{chen2025sdvlm} and
HiSpatial~\citep{liang2026hispatial} use relative depth, depth positional encoding
and metric point maps, respectively.
VLM-3R~\citep{fan2025vlm3r}, Spatial-MLLM~\citep{wu2025spatialmllm} and
VG-LLM~\citep{zheng2025vgllm} fuse visual and 3D features;
SpaceMind~\citep{zhao2025spacemind} uses camera-guided fusion, and
SpatialStack~\citep{zhang2026spatialstack} spreads geometry across LLM layers.
These studies vary the geometric source and fusion; we train the interface for the receiving VLM.

\paragraph{Learning geometry representations for reasoning.}
G$^2$VLM~\citep{hu2025g2vlm} combines reconstruction with spatial understanding,
and GAP-MLLM~\citep{zhang2026gapmllm} pretrains with sparse point-map and semantic prediction.
SpatioLM~\citep{wu2026spatiolm} uses pseudo depth and camera supervision but only RGB at
inference; GeoSR~\citep{zhang2026geosr} finds that fused VGGT tokens are often underused and masks
visual tokens during training to force their use, with QA as the only supervision.
3DThinker~\citep{chen2026think3d} aligns generated latents with VGGT features, and
CoVT~\citep{qin2025covt} learns continuous visual thoughts;
LVR~\citep{li2026lvr} and Monet~\citep{wang2026monet} reason with visual latents,
and Coconut~\citep{hao2024training} reasons in a continuous latent space.
Unlike these methods, \georeader{} supervises its geometric input through description prediction by the frozen VLM, and takes estimated pose and depth, not descriptions, at inference.

\section{Method}
\label{sec:method}

\begin{figure}[!t]
  \centering
  \includegraphics[width=\linewidth]{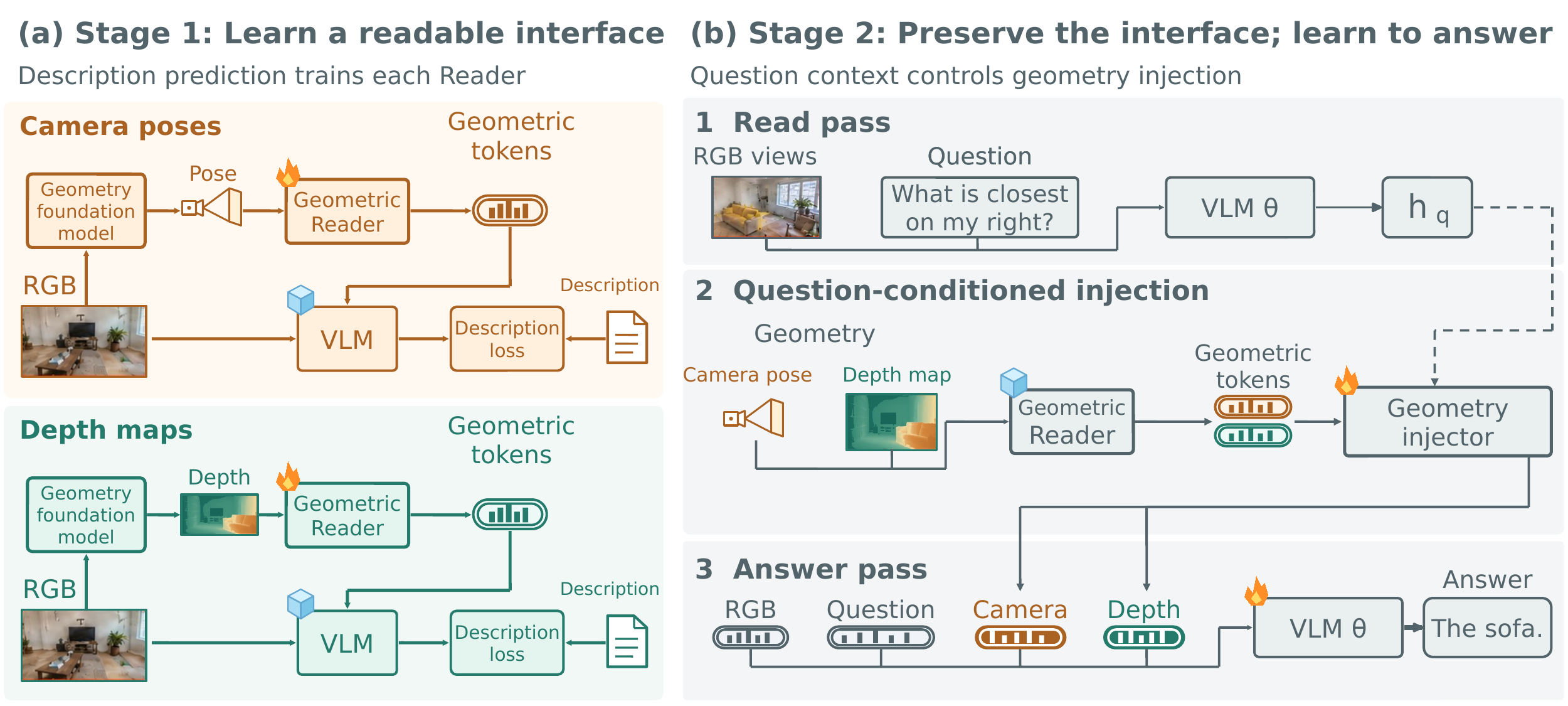}
  \caption{Overview of \georeader{}. Depth Anything 3~\citep{lin2025da3} predicts camera poses and depth maps from RGB. (a) Stage~1 trains Readers through a frozen VLM's description loss. (b) With Readers frozen, the injector uses read-pass context $h_q$ to map $Z^m$ to answer-pass tokens $\widetilde Z^m$. Flames and ice mark trainable and frozen modules; descriptions train Readers only.}
  \label{fig:method-overview}
\end{figure}

We aim to supply a pretrained VLM with geometric input that meets two separate
requirements: the geometric tokens must be \emph{readable} by the receiving model, and the model must \emph{use} them for the question at
hand. \georeader{} addresses them with two training stages
(\cref{fig:method-overview}). After the preliminaries
(\cref{sec:method-notation}), we describe how a Geometric Reader encodes
each view as a geometric token (\cref{sec:method-encoding}), how Stage~1
trains it for readability through the frozen VLM
(\cref{sec:method-readability}), and how Stage~2 injects these tokens
according to the question (\cref{sec:method-injection}).

\subsection{Preliminaries}
\label{sec:method-notation}

A VLM with parameters $\theta$ encodes the $V$ images $I=\{I_i\}_{i=1}^{V}$
into visual tokens $X_1,\ldots,X_V$, embeds the question as text tokens
$Q$, and feeds the sequence of $d$-dimensional embeddings to a language
model that predicts the answer $a$ with next-token distribution $p_\theta$.
Each view $i$ also has geometric signals $G^{m}_i$, $m\in\mathcal{M}$, here camera pose relative to view~1 and a depth map; the formulation applies to any per-view signal. We turn $G^{m}$ into additional tokens of this input sequence.

\subsection{Geometric Token Encoding}
\label{sec:method-encoding}

To feed geometry through the VLM's existing input channel, a lightweight
Geometric Reader $E_m$ maps each view of signal $m$ to one continuous input
embedding. For view~$i$, it first computes per-patch descriptors
$U^m_i\in\mathbb{R}^{N_i^m\times c_m}$ on the same patch grid as the RGB tokens;
each $c_m$-dimensional descriptor holds the patch position together with
pose-related quantities or depth statistics of that patch. A learned query $\mathbf{r}^m$ then reads only
that view's descriptors:
\begin{equation}
\begin{gathered}
    M^m_i = \mathrm{MLP}_m\bigl(\mathrm{LN}(U^m_i)\bigr),
    \qquad
    Z^m_i = \mathrm{Dec}_m\bigl(\mathbf{r}^m, M^m_i\bigr)\in\mathbb{R}^{1\times d},\\
    Z^{m} = E_m(G^{m}) = \bigl[Z^m_1;\ldots;Z^m_V\bigr]\in\mathbb{R}^{V\times d},
\end{gathered}
\end{equation}
where $\mathrm{Dec}_m$ applies two Transformer decoder blocks with
$\mathbf{r}^m$ cross-attending to $M^m_i$, then a linear projection. We call
$Z^{m}$ \emph{geometric tokens}: continuous embeddings, one per view in
input-view order. The Reader encodes geometry alone; the VLM combines it with RGB through its own attention layers. Each signal has its own Reader, so signals keep their own structure
(\cref{sec:appendix-reader-architecture}).

\subsection{Frozen-VLM Readability Training}
\label{sec:method-readability}

QA supervision alone does not specify what the geometry encodes, and feature
alignment measures similarity to an external target rather than what the
receiving VLM can read (\cref{sec:introduction}). We instead supervise each Reader through the frozen target VLM: given the
images, the geometric tokens and a fixed instruction, the VLM predicts a
geometry description $y^{m}$ of signal $m$ with teacher forcing.
For two ScanNet~\citep{dai2017scannet} views, the camera description reads \emph{``compared with image~1,
the observer in image~2 moves 0.4 meters right and 0.1 meters backward, and
turns left by about 23 degrees,''} and the depth description lists objects per
view, e.g., \emph{``guitar (left) at 1.7 meters.''} The supervised positions
$\mathcal{T}^{m}$ cover geometric content such as distances, directions and
angles; template words, predictable without geometry, are excluded. The
description loss is
\begin{equation}
\label{eq:readability}
    \mathcal{L}_{\mathrm{read}}^{m}
    = -\frac{1}{|\mathcal{T}^{m}|}\sum_{t\in\mathcal{T}^{m}} \log p_{\theta}
      \bigl(y^{m}_{t} \mid I, Z^{m}, y^{m}_{<t}\bigr),
    \qquad \theta\ \text{frozen},
\end{equation}
where gradients pass through the frozen VLM to update only $E_m$.
This trains the Reader to produce geometric tokens readable by the VLM, with descriptions used only as training targets.

\paragraph{Pair-supervised readability.}
Descriptions of a signal such as camera pose relate each view to reference
view~1. \emph{Group supervision} supervises the descriptions of all $V$ views jointly
in one example; the description of
view~$i$ can then be predicted from any view's tokens, so the loss does not
tell the Reader which token should carry that relation. \emph{Pair supervision}
removes this ambiguity by applying \cref{eq:readability} to each
reference--target pair: %
for target view $i\in\{2,\ldots,V\}$, the Reader encodes only the geometry of
views~1 and $i$, and the frozen VLM predicts view~$i$'s description
relative to view~1 from these two views, so only their tokens can explain it.
Stage~2 inputs keep all views (\cref{sec:appendix-stage1-data}).

\subsection{Question-Conditioned Geometry Injection}
\label{sec:method-injection}

\paragraph{Freezing the Readers.}
Stage~2 supervises answers, which do not explicitly require the Readers to
retain the geometric relationships specified by the Stage~1 descriptions.
Updating the Readers under this supervision may weaken their ability to
express geometry beyond what the training questions need. We therefore freeze the Readers and adapt the VLM to this fixed interface.

\paragraph{Question-conditioned readout.}
Questions differ in how much they rely on each geometric signal, so we let the
question set each signal's strength. We compute it from the final-layer hidden
state of the last valid question token, which attends to all preceding visual
and question tokens and thus summarizes the task context. Geometric
embeddings, however, enter before layer~1, and one forward pass cannot use its
own final-layer state to set earlier inputs. We therefore first run a read pass
over only the images and question to obtain this hidden state, $h_q$, before adding geometry.

\paragraph{Geometry injector.}
A geometry injector turns the frozen Reader outputs into the geometric tokens
of the answer pass. For each signal $m$, it maps $Z^m$ through an adapter
$P_m$, scales the result by a gate computed from $h_q$ with a routing head $R_m$, and adds it to the slot embeddings $B^m\in\mathbb{R}^{V\times d}$, the
learned input embeddings of the placeholder tokens at the geometric-token
positions:
\begin{equation}
    \widetilde{Z}^{m} = B^m + g_m P_m(Z^m),
    \qquad
    g_m = \mathrm{sigmoid}\bigl(R_m(h_q)\bigr).
\end{equation}
The adapter fits the frozen Reader outputs to the VLM, and
$g_m$ is the question-dependent strength of signal $m$. If a signal is missing (e.g., inter-view geometry for one image), $g_m=0$ and the slot embeddings alone mark it as absent, keeping the layout unchanged.

\paragraph{Training and inference.}
The answer pass places $\widetilde{Z}^{m}$ after the question, so the passes
encode
\begin{equation}
    S_{\mathrm{read}} = \bigl[X_1,\ldots,X_V,Q\bigr],
    \qquad
    S_{\mathrm{ans}} = \bigl[X_1,\ldots,X_V,Q,\{\widetilde{Z}^{m}\}_{m\in\mathcal{M}},a\bigr],
\end{equation}
where $a$ is the ground-truth answer in training and is generated at inference. Both passes share the VLM parameters, and we stop
gradients through the read pass to save memory and computation.
We fine-tune the language model and the geometry injector with next-token
cross-entropy on the answer and keep the Readers and the vision encoder frozen.
Since the geometric tokens come after the question, $S_{\mathrm{ans}}$ begins
with $S_{\mathrm{read}}$; at inference, the answer pass reuses the read pass's
key--value cache and computes only the geometric tokens and the answer.

\section{Experiments}
\label{sec:experiments}
\begin{table}[!t]
\caption{MindCube-Tiny~\citep{yin2025mindcube} accuracy (\%). Per API or backbone-size block, \bestkey{} marks the best and \secondbestkey{} the runner-up (only with $\geq 3$ models; base rows unshaded); $^{\dagger}$: evaluated by us; sizes in parentheses are total parameters reported by the provider. \georeader{} uses DA3 geometry and achieves the highest accuracy in four of the five backbone blocks.}
  \label{tab:mindcube-ego3d-results}
  \centering
  \small
  \setlength{\tabcolsep}{2.75pt}
  \begin{tabularx}{\linewidth}{@{}l*{3}{>{\centering\arraybackslash}X}>{\columncolor{white}[\tabcolsep][0pt]\centering\arraybackslash}X@{}}
    \filledtoprule
\tablepadabove{Method} & Rot. & Among & Around & Overall $\uparrow$ \\
      \tablegrouprule
      \rowcolor{TableGroupFill}[0pt][0pt]
      \multicolumn{5}{@{}l@{}}{\strut\tablegroup{API models}} \\
      \addlinespace[1pt]
\rule{0pt}{\dimexpr\ht\strutbox+1pt\relax}\includegraphics[height=7pt]{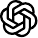}\, GPT-4.1~\citep{openai2025gpt41} & $45.5$ & \cellcolor{TableSecondFill}$44.2$ & $47.2$ & $45.1$ \\
\includegraphics[height=7pt]{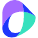}\, Doubao-Seed-1.6 (230B)~\citep{bytedance2025seed16} & \cellcolor{TableSecondFill}$87.0$ & $35.8$ & $38.0$ & $46.1$ \\
\includegraphics[height=7pt]{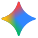}\, Gemini-2.5-Pro~\citep{gemini2025} & $84.0$ & $39.7$ & $56.8$ & $52.2$ \\
\includegraphics[height=7pt]{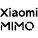}\, MiMo-V2.6-Flash (309B)$^{\dagger}$~\citep{xiaomi2026mimov26flash} & $81.0$ & $41.3$ & $64.8$ & $54.5$ \\
\includegraphics[height=7pt]{figures/model_logos/openai.pdf}\, OpenAI o3~\citep{openai2025o3} & $86.5$ & $42.7$ & \cellcolor{TableSecondFill}$66.0$ & \cellcolor{TableSecondFill}$56.6$ \\
\includegraphics[height=7pt]{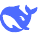}\, DeepSeek-Flash (thinking, 552B)$^{\dagger}$~\citep{deepseek2026v41flash} & \cellcolor{TableBestFill}$\mathbf{89.5}$ & \cellcolor{TableBestFill}$\mathbf{58.3}$ & \cellcolor{TableBestFill}$\mathbf{75.6}$ & \cellcolor{TableBestFill}$\mathbf{68.4}$ \\
      \tablegrouprule
      \rowcolor{TableGroupFill}[0pt][0pt]
      \multicolumn{5}{@{}l@{}}{\strut\tablegroup{Qwen2.5-VL family}} \\
      \tablegroupgap
Qwen2.5-VL-3B$^{\dagger}$~\citep{bai2025qwen} & $35.5$ & $36.2$ & $45.6$ & $38.3$ \\
HATCH-3B~\citep{oi2026hatch} & $31.0$ & $50.2$ & $65.6$ & $50.2$ \\
SenseNova-SI-3B~\citep{sensenova2025} & -- & -- & -- & $52.6$ \\
DR-MV3D-3B~\citep{choi2026drmv3d} & \cellcolor{TableSecondFill}$43.0$ & $71.3$ & $73.6$ & $66.5$ \\
3DThinker-3B~\citep{chen2026think3d} & \cellcolor{TableBestFill}$\mathbf{55.5}$ & \cellcolor{TableSecondFill}$81.8$ & \cellcolor{TableSecondFill}$75.2$ & \cellcolor{TableSecondFill}$75.2$ \\
\tablepadbelow \ours{}-3B & $40.5$ & \cellcolor{TableBestFill}$\mathbf{90.3}$ & \cellcolor{TableBestFill}$\mathbf{91.6}$ & \cellcolor{TableBestFill}$\mathbf{81.1}$ \\
      \filledmidrule
\tablepadabove{Qwen2.5-VL-7B$^{\dagger}$~\citep{bai2025qwen}} & $35.0$ & $33.2$ & $38.4$ & $34.8$ \\
SenseNova-SI-7B~\citep{sensenova2025} & -- & -- & -- & $54.7$ \\
3DThinker-7B~\citep{chen2026think3d} & \cellcolor{TableSecondFill}$55.0$ & \cellcolor{TableSecondFill}$83.0$ & \cellcolor{TableSecondFill}$76.0$ & \cellcolor{TableSecondFill}$76.0$ \\
\ours{}-7B & \cellcolor{TableBestFill}$\mathbf{90.0}$ & \cellcolor{TableBestFill}$\mathbf{94.7}$ & \cellcolor{TableBestFill}$\mathbf{92.4}$ & \cellcolor{TableBestFill}$\mathbf{93.2}$ \\
      \tablegrouprule
      \rowcolor{TableGroupFill}[0pt][0pt]
      \multicolumn{5}{@{}l@{}}{\strut\tablegroup{Qwen3-VL family}} \\
      \tablegroupgap
Qwen3-VL-2B$^{\dagger}$~\citep{bai2025qwen3vl} & $29.5$ & $33.3$ & $47.2$ & $35.9$ \\
Spatial-TTT-2B~\citep{liu2026spatialttt} & \cellcolor{TableSecondFill}$55.5$ & $74.0$ & \cellcolor{TableSecondFill}$89.8$ & \cellcolor{TableSecondFill}$76.2$ \\
3DThinker-2B~\citep{chen2026think3d} & $53.0$ & \cellcolor{TableSecondFill}$83.2$ & $78.4$ & \cellcolor{TableSecondFill}$76.2$ \\
\tablepadbelow \ours{}-2B & \cellcolor{TableBestFill}$\mathbf{59.5}$ & \cellcolor{TableBestFill}$\mathbf{93.3}$ & \cellcolor{TableBestFill}$\mathbf{91.6}$ & \cellcolor{TableBestFill}$\mathbf{86.5}$ \\
      \filledmidrule
\tablepadabove{Qwen3-VL-4B$^{\dagger}$~\citep{bai2025qwen3vl}} & $25.0$ & $25.3$ & $43.6$ & $29.6$ \\
3DThinker-4B~\citep{chen2026think3d} & \cellcolor{TableSecondFill}$55.0$ & \cellcolor{TableSecondFill}$81.7$ & \cellcolor{TableSecondFill}$78.4$ & \cellcolor{TableSecondFill}$75.8$ \\
\tablepadbelow \ours{}-4B & \cellcolor{TableBestFill}$\mathbf{94.5}$ & \cellcolor{TableBestFill}$\mathbf{96.2}$ & \cellcolor{TableBestFill}$\mathbf{89.6}$ & \cellcolor{TableBestFill}$\mathbf{94.3}$ \\
      \filledmidrule
\tablepadabove{Qwen3-VL-8B$^{\dagger}$~\citep{bai2025qwen3vl}} & $42.0$ & $32.3$ & $51.2$ & $38.7$ \\
SenseNova-SI-8B~\citep{sensenova2025} & \cellcolor{TableSecondFill}$79.5$ & $73.2$ & $70.4$ & $73.8$ \\
3DThinker-8B~\citep{chen2026think3d} & $56.5$ & \cellcolor{TableSecondFill}$82.5$ & \cellcolor{TableSecondFill}$78.8$ & $76.7$ \\
GAMSI-8B~\citep{zheng2026gamsi} & -- & -- & -- & \cellcolor{TableBestFill}$\mathbf{94.1}$ \\
\tablepadbelow \ours{}-8B & \cellcolor{TableBestFill}$\mathbf{91.5}$ & \cellcolor{TableBestFill}$\mathbf{93.3}$ & \cellcolor{TableBestFill}$\mathbf{95.2}$ & \cellcolor{TableSecondFill}$93.4$ \\
    \filledbottomrule
  \end{tabularx}
\end{table}

We first compare \georeader{} with existing models and vanilla SFT (the same
QA fine-tuning without geometry; \cref{sec:spatial-performance}). Next, we test
Reader training and freezing (\cref{sec:readability-evaluation}), then injection
designs and whether answers use geometry (\cref{sec:utilization-evaluation}).
Finally, we compare geometric tokens with text descriptions and present
qualitative examples (\cref{sec:interface-tradeoffs}).

\subsection{Experimental Setup}
\label{sec:experimental-setup}
For training, Stage~1 uses about 65K geometry descriptions from 1,550 ScanNet
and ScanNet++ training scenes (\cref{sec:appendix-stage1-data}), and Stage~2
uses about 135K multi-source QA examples for main models and 10K MindCube
examples for ablations, none from test sets
(\cref{sec:appendix-training-mixture}).
We evaluate on MindCube-Tiny~\citep{yin2025mindcube},
VSI-Bench~\citep{yang2024vsi}, SPBench~\citep{li2025spatialladder},
CV-Bench~\citep{tong2024cambrian}, SPAR-Bench~\citep{zhang2025spar},
ViewSpatial~\citep{li2025viewspatial} and MMSI-Bench~\citep{yang2025mmsi}
(\cref{sec:appendix-eval-protocol}). Main models are trained and evaluated with
geometry predicted from RGB by Depth Anything 3~\citep{lin2025da3} (DA3;
errors in \cref{tab:da3-geometry-error}), and ablations use annotated
geometry to exclude estimation errors.
We build \georeader{} on 3B/7B Qwen2.5-VL~\citep{bai2025qwen}, 4B/8B
Qwen3-VL~\citep{bai2025qwen3vl} and 4B/9B Qwen3.5~\citep{qwen2026qwen35}
with one recipe: Readers train for 10 epochs with the VLM frozen, and Stage~2
fine-tunes for 2 epochs on the multi-source data or 10 on MindCube at learning
rate $2\times10^{-5}$, with greedy decoding
(full training settings in \cref{sec:appendix-training-settings}).

\subsection{Main Results}
\label{sec:spatial-performance}
\paragraph{Comparison with existing models.}
\cref{tab:mindcube-ego3d-results,tab:generalization-results}
compare \georeader{} (DA3 geometry) with existing models.
On MindCube-Tiny, \georeader{} exceeds 3DThinker at every backbone size. On six more benchmarks, \georeader{} with
Qwen2.5-VL-3B averages $63.6$ versus $60.4$ for 3DThinker-3B, and up to
$72.5$ with Qwen3.5-9B. Over vanilla SFT, geometry helps metric-geometry questions and leaves others unchanged (\cref{tab:question-type-gains}),
with gains up to $+14.3$ on SPBench-SI relative distance and $+13.5$ on
MMSI camera motion (\cref{sec:appendix-rgb-only-comparison}).
Gains thus come from geometry, not QA fine-tuning.

\begin{table}[!t]
\caption{Six-benchmark accuracy (\%); Avg.\ is the mean. Per backbone-size block, \bestkey{} marks the best and \secondbestkey{} the runner-up (only with $\geq 3$ models; base rows unshaded); $^{\dagger}$: evaluated by us; $^{*}$: combined from two sources (\cref{sec:appendix-baseline-sources}). \georeader{} has the highest average in every block.}
  \label{tab:generalization-results}
  \centering
  \small
  \setlength{\tabcolsep}{2pt}
    \begin{tabularx}{\linewidth}{@{}l>{\hsize=0.92\hsize\centering\arraybackslash}X>{\hsize=1.1\hsize\centering\arraybackslash}X>{\hsize=0.92\hsize\centering\arraybackslash}X>{\hsize=0.92\hsize\centering\arraybackslash}X>{\hsize=1.3\hsize\centering\arraybackslash}X>{\hsize=0.92\hsize\centering\arraybackslash}X>{\columncolor{white}[\tabcolsep][0pt]\hsize=0.92\hsize\centering\arraybackslash}X@{}}
      \filledtoprule
\tablepadabove{Method} & VSI & SPBench & CV & SPAR & ViewSpatial & MMSI & Avg. $\uparrow$ \\
      \tablegrouprule
      \rowcolor{TableGroupFill}[0pt][0pt]
      \multicolumn{8}{@{}l@{}}{\strut\tablegroup{Qwen2.5-VL family}} \\
      \tablegroupgap
Qwen2.5-VL-3B$^{\dagger}$~\citep{bai2025qwen} & $27.3$ & $39.5$ & $70.5$ & $27.9$ & $36.1$ & $25.5$ & $37.8$ \\
Spatial-MLLM-4B~\citep{wu2025spatialmllm} & $47.3$ & $48.4$ & $73.8$ & $35.1$ & $43.6$ & $31.5$ & $46.6$ \\
SpatialLadder-3B~\citep{li2025spatialladder} & $45.7$ & \cellcolor{TableBestFill}$\mathbf{70.6}$ & $73.7$ & $34.4$ & $44.2$ & $29.2$ & $49.6$ \\
3DThinker-3B~\citep{chen2026think3d} & \cellcolor{TableBestFill}$\mathbf{59.1}$ & $60.2$ & \cellcolor{TableSecondFill}$78.4$ & \cellcolor{TableSecondFill}$58.2$ & \cellcolor{TableSecondFill}$64.7$ & \cellcolor{TableBestFill}$\mathbf{41.9}$ & \cellcolor{TableSecondFill}$60.4$ \\
\tablepadbelow \ours{}-3B & \cellcolor{TableSecondFill}$56.1$ & \cellcolor{TableSecondFill}$70.0$ & \cellcolor{TableBestFill}$\mathbf{83.7}$ & \cellcolor{TableBestFill}$\mathbf{59.1}$ & \cellcolor{TableBestFill}$\mathbf{71.4}$ & \cellcolor{TableSecondFill}$41.2$ & \cellcolor{TableBestFill}$\mathbf{63.6}$ \\
      \filledmidrule
\tablepadabove{Qwen2.5-VL-7B$^{\dagger}$~\citep{bai2025qwen}} & $29.9$ & $40.8$ & $77.9$ & $33.2$ & $34.7$ & $26.8$ & $40.5$ \\
SpaceR-7B~\citep{ouyang2025spacer} & $44.5$ & $54.0$ & $75.3$ & $37.1$ & $45.5$ & $28.8$ & $47.5$ \\
ViLaSR-7B~\citep{wu2025vilasr} & $45.4$ & $53.9$ & $77.1$ & $37.6$ & $46.1$ & $30.2$ & $48.4$ \\
Video-R1~\citep{feng2025videor1} & $33.4$ & $42.8$ & $69.6$ & $31.5$ & $36.1$ & $29.4$ & $40.5$ \\
3DThinker-7B~\citep{chen2026think3d} & \cellcolor{TableBestFill}$\mathbf{63.7}$ & \cellcolor{TableSecondFill}$68.3$ & \cellcolor{TableSecondFill}$81.1$ & \cellcolor{TableBestFill}$\mathbf{63.3}$ & \cellcolor{TableSecondFill}$68.6$ & \cellcolor{TableSecondFill}$43.3$ & \cellcolor{TableSecondFill}$64.7$ \\
\ours{}-7B & \cellcolor{TableSecondFill}$58.0$ & \cellcolor{TableBestFill}$\mathbf{72.3}$ & \cellcolor{TableBestFill}$\mathbf{84.4}$ & \cellcolor{TableSecondFill}$61.4$ & \cellcolor{TableBestFill}$\mathbf{75.6}$ & \cellcolor{TableBestFill}$\mathbf{45.5}$ & \cellcolor{TableBestFill}$\mathbf{66.2}$ \\
      \tablegrouprule
      \rowcolor{TableGroupFill}[0pt][0pt]
      \multicolumn{8}{@{}l@{}}{\strut\tablegroup{Qwen3-VL family}} \\
      \tablegroupgap
Qwen3-VL-4B$^{\dagger}$~\citep{bai2025qwen3vl} & $55.1$ & $58.2$ & $84.4$ & $33.4$ & $38.6$ & $27.2$ & $49.5$ \\
\tablepadbelow \ours{}-4B & \cellcolor{TableBestFill}$\mathbf{64.8}$ & \cellcolor{TableBestFill}$\mathbf{78.9}$ & \cellcolor{TableBestFill}$\mathbf{85.6}$ & \cellcolor{TableBestFill}$\mathbf{65.3}$ & \cellcolor{TableBestFill}$\mathbf{77.4}$ & \cellcolor{TableBestFill}$\mathbf{43.6}$ & \cellcolor{TableBestFill}$\mathbf{69.3}$ \\
      \filledmidrule
\tablepadabove{Qwen3-VL-8B$^{\dagger}$~\citep{bai2025qwen3vl}} & $55.9$ & $57.8$ & $85.4$ & $38.2$ & $40.4$ & $28.9$ & $51.1$ \\
SenseNova-SI-8B$^{*}$~\citep{sensenova2025} & \cellcolor{TableSecondFill}$64.8$ & \cellcolor{TableSecondFill}$65.8$ & $83.4$ & \cellcolor{TableSecondFill}$41.1$ & \cellcolor{TableSecondFill}$51.2$ & \cellcolor{TableSecondFill}$38.1$ & \cellcolor{TableSecondFill}$57.4$ \\
\ours{}-8B & \cellcolor{TableBestFill}$\mathbf{66.2}$ & \cellcolor{TableBestFill}$\mathbf{76.7}$ & \cellcolor{TableBestFill}$\mathbf{86.5}$ & \cellcolor{TableBestFill}$\mathbf{65.7}$ & \cellcolor{TableBestFill}$\mathbf{78.5}$ & \cellcolor{TableBestFill}$\mathbf{46.6}$ & \cellcolor{TableBestFill}$\mathbf{70.0}$ \\
      \tablegrouprule
      \rowcolor{TableGroupFill}[0pt][0pt]
      \multicolumn{8}{@{}l@{}}{\strut\tablegroup{Qwen3.5 family}} \\
      \addlinespace[1pt]
{\setlength{\belowrulesep}{1pt}\tablepadabove{Qwen3.5-4B$^{\dagger}$~\citep{qwen2026qwen35}}} & $50.0$ & $35.5$ & \cellcolor{TableBestFill}$\mathbf{84.9}$ & $34.4$ & $34.5$ & $28.1$ & $44.6$ \\
\tablepadbelow \ours{}-4B & \cellcolor{TableBestFill}$\mathbf{66.7}$ & \cellcolor{TableBestFill}$\mathbf{81.5}$ & $81.8$ & \cellcolor{TableBestFill}$\mathbf{68.7}$ & \cellcolor{TableBestFill}$\mathbf{77.7}$ & \cellcolor{TableBestFill}$\mathbf{45.4}$ & \cellcolor{TableBestFill}$\mathbf{70.3}$ \\
      \filledmidrule
\tablepadabove{Qwen3.5-9B$^{\dagger}$~\citep{qwen2026qwen35}} & $54.1$ & $50.7$ & $85.9$ & $42.1$ & $38.2$ & $32.0$ & $50.5$ \\
\tablepadbelow \ours{}-9B & \cellcolor{TableBestFill}$\mathbf{68.6}$ & \cellcolor{TableBestFill}$\mathbf{79.5}$ & \cellcolor{TableBestFill}$\mathbf{87.7}$ & \cellcolor{TableBestFill}$\mathbf{71.0}$ & \cellcolor{TableBestFill}$\mathbf{81.3}$ & \cellcolor{TableBestFill}$\mathbf{47.1}$ & \cellcolor{TableBestFill}$\mathbf{72.5}$ \\
      \filledbottomrule
    \end{tabularx}%
\end{table}

\begin{table}[!t]
\caption{Accuracy (\%) of vanilla SFT and \georeader{} by question type.
Both use Qwen2.5-VL-3B~\citep{bai2025qwen} with matched data and settings; the 59 subtasks of the
six benchmarks are grouped by whether they require metric geometry
(subtask counts in parentheses), weighted by question count. Geometry helps
questions requiring metric geometry without hurting the others.}
\label{tab:question-type-gains}
\centering\small
\begin{tabularx}{\linewidth}{@{}l*{3}{>{\centering\arraybackslash}X}>{\columncolor{TableSecondFill}[\tabcolsep][0pt]\centering\arraybackslash}X@{}}
\filledtoprule
\tablepadbelow \tablepadabove{Question type} & Questions & Vanilla & \ours{} & $\Delta$ \\
\filledmidrule
\tablepadabove{Object and room size (4)} & $1{,}862$ & $65.2$ & $70.0$ & $\mathbf{+4.8}$ \\
Camera motion (4) & $967$ & $39.3$ & $43.7$ & $\mathbf{+4.4}$ \\
\tablepadbelow Distance and depth (19) & $6{,}613$ & $56.8$ & $59.0$ & $\mathbf{+2.2}$ \\
\filledmidrule
\tablepadabove{Metric-geometry questions (27)} & $9{,}442$ & $56.7$ & $59.6$ & $\mathbf{+2.9}$ \\
\tablepadbelow All other questions (32) & $13{,}577$ & $67.3$ & $67.5$ & $\mathbf{+0.3}$ \\
\filledbottomrule
\end{tabularx}
\end{table}

\subsection{Learning VLM-Readable Geometry}
\label{sec:readability-evaluation}
Ablations use Qwen2.5-VL-3B and the 1,050-question MindCube-Tiny test set.
Stage~1 alone, without QA training, scores $41.4\%$ (\cref{tab:stage-results})
and QA training alone $54.7\%$ (\cref{tab:alignment-ablation}, Panel A),
whereas combining them achieves $89.2\%$. This shows that both stages are needed.

\paragraph{Reader pretraining.}
Panel A of \cref{tab:alignment-ablation} compares the two routes of \cref{sec:introduction} with ours. Readability pretraining achieves $76.4\%$, above QA-only
training of a random Reader ($54.7\%$) and %
feature alignment ($62.7\%$; released 3DThinker checkpoint~\citep{chen2026think3d}, same backbone and questions).

\paragraph{Pair versus group supervision.}
Panel C of \cref{tab:text-cot-comparison} compares the
supervisions of \cref{sec:method-readability} with frozen Readers. Pair supervision, which removes
the view ambiguity of group supervision (all views in one example), improves
accuracy from $77.8\%$ to $89.2\%$ (\cref{sec:appendix-text-interface-protocol}).

\afterpage{%
\begin{table}[!t]
\centering
\begin{minipage}[t]{0.40\linewidth}
\vspace{0pt}
\caption{MindCube-Tiny~\citep{yin2025mindcube} accuracy (\%). A: QA-tuned Readers; B: complete interfaces. Readability pretraining leads A; \georeader{} leads B with its Reader frozen.}
\label{tab:alignment-ablation}
\end{minipage}\hfill
\begin{minipage}[t]{0.57\linewidth}
\vspace{0pt}
\caption{MindCube-Tiny~\citep{yin2025mindcube} accuracy (\%), generated tokens and latency (s) with annotated geometry; Text-CoT generates descriptions before the answer. Pair-supervised \georeader{} is most accurate, with as few generated tokens as vanilla SFT.}
\label{tab:text-cot-comparison}
\end{minipage}\par
\begin{minipage}[t]{0.40\linewidth}
\vspace{0pt}
\centering\small
\setlength{\tabcolsep}{2pt}
\setlength{\extrarowheight}{0.63pt}
\begin{tabular*}{\linewidth}{@{\extracolsep{\fill}}lc@{}}
\toprule
Geometric interface & Overall $\uparrow$ \\
\tablegrouprule
\rowcolor{TableGroupFill}[0pt][0pt]
\multicolumn{2}{@{}l@{}}{\strut\tablegroup{A. Reader pretraining, then tuned in QA}} \\
\tablegroupgap
No Stage~1 (QA loss only) & $54.7$ \\
Feat.\ align.\ (released) & $62.7$ \\
Readability pretraining & $76.4$ \\
\tablegrouprule
\rowcolor{TableGroupFill}[0pt][0pt]
\multicolumn{2}{@{}l@{}}{\strut\tablegroup{B. Complete interfaces}} \\
\tablegroupgap
VLM-3R-style (VGGT) & $86.1$ \\
SpaceMind-style (VGGT) & $81.1$ \\
Per-token additive (pose) & $73.8$ \\
\ours{} (frozen Reader) & $\mathbf{89.2}$ \\
\bottomrule
\end{tabular*}
\end{minipage}\hfill
\begin{minipage}[t]{0.57\linewidth}
\vspace{0pt}
\centering\small
\setlength{\tabcolsep}{2pt}
\begin{tabular*}{\linewidth}{@{\extracolsep{\fill}}lccc@{}}
\toprule
Recipe & Acc.\ $\uparrow$ & Tokens & Latency (s) $\downarrow$ \\
\tablegrouprule
\rowcolor{TableGroupFill}[0pt][0pt]
\multicolumn{4}{@{}l@{}}{\strut\tablegroup{A. No geometry}} \\
\tablegroupgap
Vanilla SFT & $68.6$ & $16.2$ & $0.5$ \\
\tablegrouprule
\rowcolor{TableGroupFill}[0pt][0pt]
\multicolumn{4}{@{}l@{}}{\strut\tablegroup{B. Text geometry}} \\
\tablegroupgap
Joint description + QA SFT & $67.0$ & $16.3$ & $0.5$ \\
Generated Text-CoT & $72.9$ & $119.3$ & $2.5$ \\
Descriptions in prompt & $81.0$ & $16.3$ & $0.5$ \\
\tablegrouprule
\rowcolor{TableGroupFill}[0pt][0pt]
\multicolumn{4}{@{}l@{}}{\strut\tablegroup{C. Continuous geometry}} \\
\tablegroupgap
\ours{} (group) & $77.8$ & $16.2$ & $0.5$ \\
\ours{} (pair) & $\mathbf{89.2}$ & $16.2$ & $0.5$ \\
\bottomrule
\end{tabular*}
\end{minipage}
\end{table}

\begin{table}[!t]
\centering
\begin{minipage}[t]{0.32\linewidth}
\vspace{0pt}
\caption{Accuracy (\%) per\protect\\ stage. Stage~1 alone is not\protect\\ enough; both are needed.}
\label{tab:stage-results}
\centering\small
\setlength{\tabcolsep}{3pt}
\begin{tabular*}{\linewidth}{@{\extracolsep{\fill}}lc@{}}
\toprule
Configuration & QA $\uparrow$ \\
\midrule
Vanilla SFT & $68.6$ \\
Stage~1 only & $41.4$ \\
Stage~1 + Stage~2 & $\mathbf{89.2}$ \\
\bottomrule
\end{tabular*}

\end{minipage}\hfill
\begin{minipage}[t]{0.32\linewidth}
\vspace{0pt}
\caption{Accuracy (\%) per\protect\\ injection design. The geometry\protect\\ injector scores highest.}
\label{tab:read-then-inject-ablation}
\centering\small
\setlength{\tabcolsep}{3pt}
\begin{tabular*}{\linewidth}{@{\extracolsep{\fill}}lc@{}}
\toprule
Injection design & QA $\uparrow$ \\
\midrule
Direct & $78.6$ \\
Learned fixed scaling & $68.5$ \\
Geometry injector & $\mathbf{89.2}$ \\
\bottomrule
\end{tabular*}
\end{minipage}\hfill
\begin{minipage}[t]{0.32\linewidth}
\vspace{0pt}
\caption{Accuracy (\%) per\protect\\ geometry intervention. Only\protect\\ \georeader{} drops markedly.}
\label{tab:token-use-ablation}
\centering\small
\setlength{\tabcolsep}{1.5pt}
\begin{tabular*}{\linewidth}{@{\extracolsep{\fill}}lcc@{}}
\toprule
Geometry & \ours{} & Feat.\ align. \\
\midrule
Shuffled & $63.1$ & $61.5$ \\
Zero & $77.2$ & $62.9$ \\
Correct & $\mathbf{89.2}$ & $62.7$ \\
\bottomrule
\end{tabular*}
\end{minipage}
\end{table}

\begin{figure}[!t]
\input{figures/token_attention_placeholder}
\end{figure}
}
\paragraph{Freezing the Readers.}
In Panel A of \cref{tab:alignment-ablation}, we tune the Reader in QA training.
Freezing the same pair-supervised Stage~1 Reader (\georeader{}, Panel B)
improves accuracy from $76.4\%$ to $89.2\%$. Because QA answers do not require the
Reader to keep Stage~1 relationships, we freeze it.

\afterpage{%
\begin{figure}[!t]
  \centering
  \includegraphics[width=\linewidth]{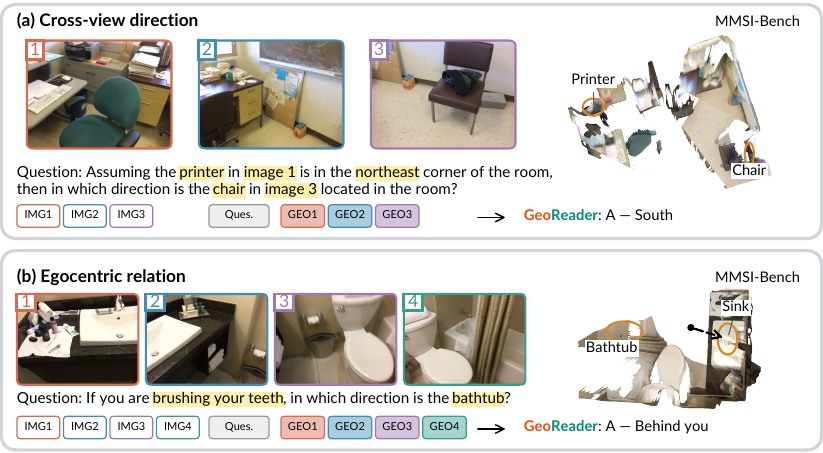}
\caption{Two MMSI-Bench~\citep{yang2025mmsi} questions \georeader{} answers correctly with DA3~\citep{lin2025da3} geometry. Right: predicted-depth point clouds, queried objects circled.}
  \label{fig:qualitative-results}
\end{figure}

}
\paragraph{Alternative interfaces.}
In Panel B of \cref{tab:alignment-ablation}, we test complete direct-injection interfaces as in Panel A. Adapted VLM-3R- and SpaceMind-style fusion on dense VGGT features achieve $86.1\%$ and $81.1\%$ (\cref{sec:appendix-adapted-fusion}), and adding annotated pose descriptors to image-token embeddings $73.8\%$, below \ours{}'s $89.2\%$ on the same geometry. This suggests that Readers suit the VLM better.

\subsection{Using Geometry for the Question}
\label{sec:utilization-evaluation}
\paragraph{Injection design.}
In \cref{tab:read-then-inject-ablation}, we replace the geometry injector with
direct injection ($78.6\%$) or a learned fixed scale ($68.5\%$); neither
matches the full injector ($89.2\%$; \cref{sec:appendix-gate-questions}).

\paragraph{Geometry interventions.}
With the checkpoint, RGB and questions fixed, zeroing \georeader{}'s geometric
residual, which keeps the slot embeddings, lowers accuracy from $89.2\%$ to
$77.2\%$, and shuffling poses across views drops it to $63.1\%$
(\cref{tab:token-use-ablation}; \cref{sec:appendix-content-protocol}). Its
answers thus depend on both the geometric content and its match to the views.
In contrast, for feature alignment (Feat.\ align.), zeroing the latents of
the same released checkpoint or swapping them across questions keeps accuracy
near $62.7\%$ ($62.9\%$ and $61.5\%$), indicating that its answers barely depend on the latents,
even though they are trained to match VGGT~\citep{wang2025vggt} features.

\paragraph{Answer-token attention.}
On a MindCube-Tiny question that both models answer correctly, we find that
\georeader{}'s answer tokens attend more to geometric tokens across layers than
3DThinker's attend to its latents (\cref{fig:token-attention};
\cref{sec:appendix-answer-attention}), consistent with the interventions.

\subsection{Interface Trade-offs and Qualitative Analysis}
\label{sec:interface-tradeoffs}
\paragraph{Text versus continuous geometry.}
As extra SFT targets, descriptions score $67.0\%$, below vanilla SFT
($68.6\%$; \cref{tab:text-cot-comparison}). Generating them at inference yields $72.9\%$ and supplying them in the prompt
$81.0\%$, whereas \georeader{}, using them only as Reader targets, achieves $89.2\%$. Descriptions thus help more as Reader supervision than as text. With precomputed geometry, \georeader{} generates $16.2$
tokens in $0.5$~s, versus $119.3$ and $2.5$~s when generating
them (\cref{sec:appendix-text-interface-protocol}).

\paragraph{Qualitative examples.}
Both questions in \cref{fig:qualitative-results} relate objects in
different views (printer/chair in views~1/3; sink/bathtub in views~1/4).
\cref{sec:appendix-predicted-examples,sec:appendix-failure-cases} give more
successes and failures.

\FloatBarrier
\section{Conclusion}
\label{sec:conclusion}

We presented \georeader{}, which converts geometry into a form that a frozen VLM
can read before the VLM learns to use it. Geometric Readers learn through
geometry-description prediction by the frozen VLM, then stay frozen while the
VLM and a question-conditioned geometry injector learn spatial QA;
descriptions serve only as training targets. Zeroing the aligned latents of a
released feature-alignment model leaves its accuracy unchanged, whereas removing \georeader{}'s geometry lowers it. With geometry predicted from RGB, \georeader{}
reaches $94.3$ on MindCube-Tiny with Qwen3-VL-4B and a six-benchmark average
of $72.5$ with Qwen3.5-9B. Capturing geometry does not ensure its use; the representation
that helps most is the one the receiving model can read.

\section*{AI Use Statement}

We used generative AI tools to assist with implementing training code,
generating training data and figures, and manuscript organization and editing.
The authors manually reviewed the AI-assisted training code
and the generated training data. The authors take responsibility for the
final content of this work, including its text, claims, code, data, and figures.

\section*{Reproducibility Statement}

\cref{sec:method} describes the model and training objectives, and
\cref{sec:appendix-reader-architecture} details the Geometric Reader
architecture. The training-data composition and split boundaries are described
in \cref{sec:appendix-training-mixture}, and geometry sources in
\cref{sec:appendix-geometry-inputs}. %
Training settings are provided in \cref{sec:appendix-training-settings}, and
the injection designs and text-geometry recipes in
\cref{sec:appendix-gate-questions,sec:appendix-text-interface-protocol}.
\cref{sec:appendix-eval-protocol,sec:appendix-content-protocol} document evaluation settings and
geometry-content interventions, respectively.

\section*{Ethics Statement}

This work uses publicly available datasets under their respective licenses or
terms (\cref{sec:appendix-public-resources}). Predicted geometry and answers can be wrong, e.g., under reference-frame confusion.
Benchmark performance does not establish reliability for safety-critical
applications such as navigation or robotic control; such use requires
application-specific validation and safeguards.

\FloatBarrier

\bibliography{iclr2027_conference}
\bibliographystyle{iclr2027_conference}

\clearpage
\appendix
\section*{Supplementary Material}
\noindent\hspace*{-14pt}%
\begin{tikzpicture}
\node[fill=TitleCardFill,rounded corners=14pt,inner xsep=14pt,inner ysep=14pt]{%
\begin{minipage}{\linewidth}
{\large\headingfont\color{GeoReaderWordmarkGreen} Contents\par}\vspace{8pt}
\begingroup
\fontsize{10}{15}\fontencoding{T1}\fontfamily{lato-TLF}\selectfont
\def\normalfont{\fontencoding{T1}\fontfamily{lato-TLF}\fontseries{m}\fontshape{n}\selectfont}
\setlength{\parskip}{0pt}
\makeatletter
\renewcommand*\l@subsection[2]{\addvspace{7pt}{\bfseries\@dottedtocline{2}{0em}{2.3em}{#1}{#2}}}
\renewcommand*\l@subsubsection{\@dottedtocline{3}{1.6em}{2.6em}}
\@starttoc{toc}
\makeatother
\endgroup
\end{minipage}};
\end{tikzpicture}\hspace*{-14pt}\par

\suppressfloats[t]

\section{Appendix}
\label{sec:appendix}
\addtocontents{toc}{\protect\setcounter{tocdepth}{3}}

\subsection{Training Data and Geometry Sources}
\label{sec:appendix-training-data}
This section describes the data and geometry used in
\cref{sec:experimental-setup}. \Cref{sec:appendix-stage1-data} covers the
Stage~1 description data (window corpus and MindCube pairs),
\cref{sec:appendix-training-mixture} the Stage~2 QA data and test-set
exclusion, and \cref{sec:appendix-geometry-inputs} the predicted and annotated
geometry.

\begin{figure}[!t]
\centering
\includegraphics[width=\linewidth]{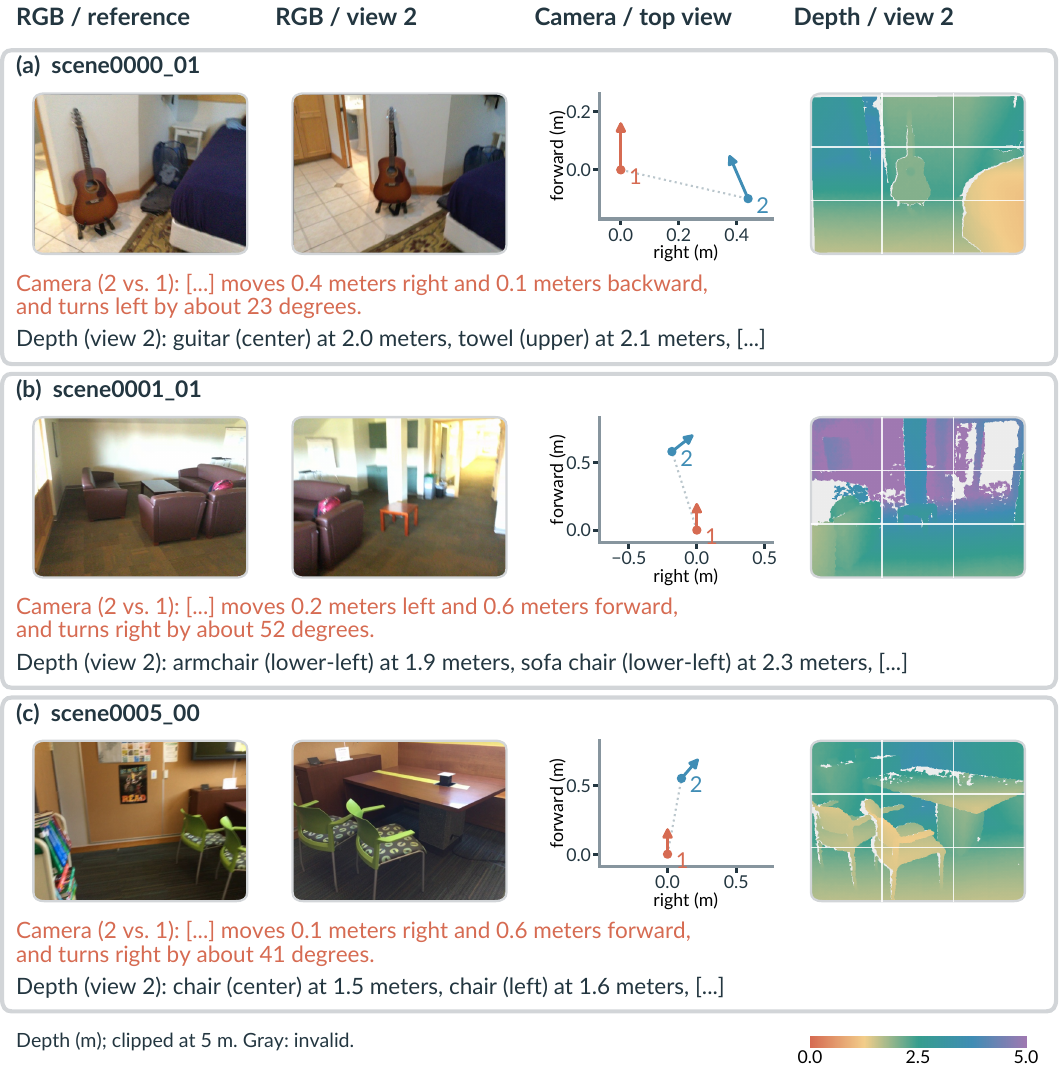}
\caption{Three Stage~1 training examples from ScanNet~\citep{dai2017scannet}: RGB pairs, camera poses in the first camera's reference frame, and view~2 depth
with nine-grid regions. Description excerpts preserve original wording and values;
[\ldots] marks omissions. Depth colors are clipped at 5\,m; gray is invalid.}
\label{fig:stage1-training-data}
\end{figure}

\subsubsection{Stage 1: Geometry-Description Data}
\label{sec:appendix-stage1-data}
\paragraph{Window corpus.}
The window-based Stage~1 corpus is built from training scenes in
ScanNet~\citep{dai2017scannet} and ScanNet++~\citep{yeshwanth2023scannetpp}.
It contains approximately 29.5K Camera and 35.4K Depth
description records, drawn from a shared pool of 1,550 scenes.
The two signals share the same windows. Camera windows contain 2, 4, 8, 16 or 32 views;
Depth windows additionally include single images.

Camera targets are constructed from annotated camera poses and describe
planar translation in meters and heading changes relative to the first view.
For each window, we adopt the first camera's coordinate frame as the reference
world frame and transform all camera poses into it. The first camera therefore
has identity pose and lies at the origin; the $x$ and $z$ axes point right and
forward relative to that view.
Depth targets describe object positions using a $3\times3$ image grid
(upper-left, upper, upper-right, left, center, right, lower-left, lower,
and lower-right), together with distances in meters. We retain up to five
object entries per view from projected instance geometry and annotations.
Single-view windows are omitted from Camera training; windows without any
retained object entries are omitted from Depth training.
Scenes in the evaluation benchmarks are excluded before sampling windows.
\cref{fig:stage1-training-data} shows three windows with their geometry and description excerpts.
The nine-grid regions define the textual position labels; distinct from the
finer geometry patch grid.

\paragraph{MindCube pairs.}
The description construction for MindCube~\citep{yin2025mindcube} is separate
from the window corpus. It uses 2,072 training image groups, yielding 3,375
reference--target pairs for pair-supervised Camera Reader pretraining
(\cref{eq:readability}); \cref{sec:appendix-text-interface-protocol} gives the
per-epoch example counts. A four-view group yields three independent pairs
$(1,2)$, $(1,3)$ and $(1,4)$, not all six possible pairs. Each pair
retains only its two images, the corresponding camera geometry and one
relative-pose description, with the reference coordinate frame and source
camera geometry preserved; pairs repeated across source groups are retained.
The excluded views and their geometry are absent from a pair's forward pass,
rather than merely excluded from the loss, and different pairs share model
weights but no attention context. Both group and pair supervision compute the
loss only at geometric-content positions (\cref{sec:method-readability}).
Pairs use training groups only; QA inputs keep all views.

\begin{figure}[!t]
\centering
\includegraphics[width=\linewidth]{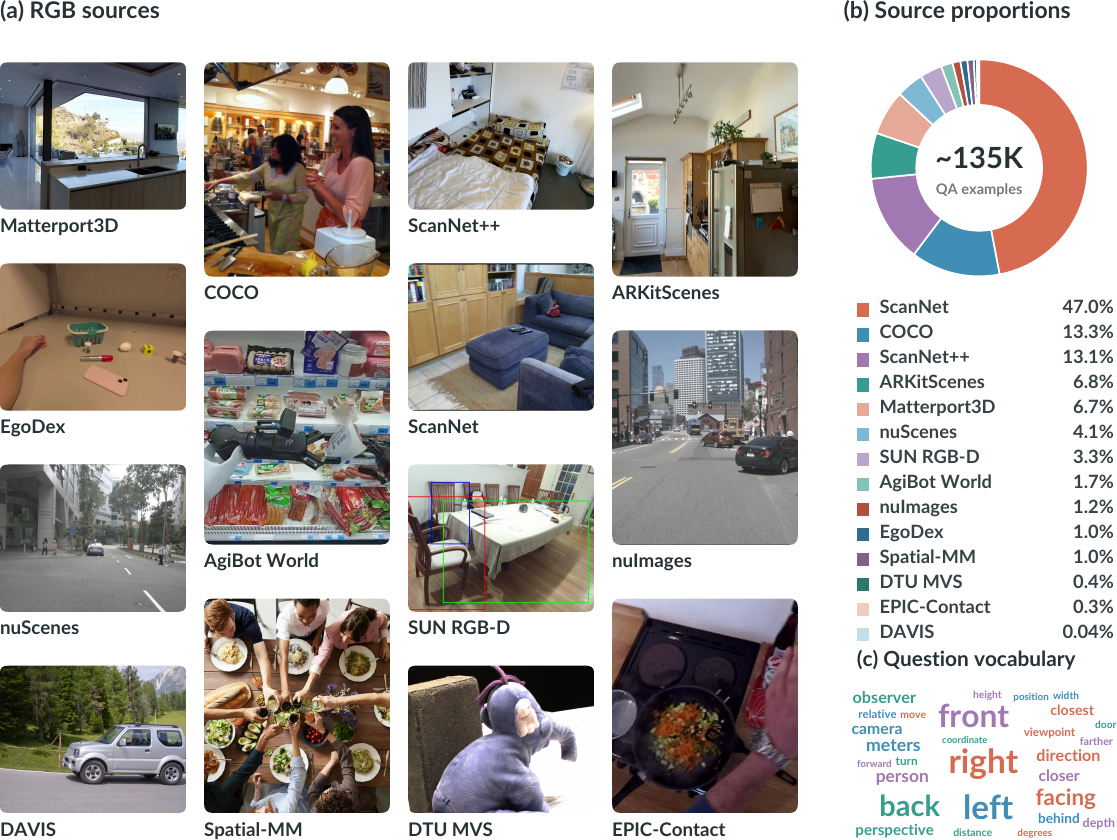}
\caption{The approximately 135K-example Stage~2 QA collection. Left: cropped RGB examples; image sizes do not encode frequency.
Right: QA shares by RGB source and selected spatial terms sized by question-stem
frequency. All $14$ RGB sources are shown.}
\label{fig:training-data-pie}
\end{figure}

\subsubsection{Stage 2: QA Data and Test-Set Exclusion}
\label{sec:appendix-training-mixture}
The multi-source QA collection has approximately 135K examples and is separate
from the 10K MindCube QA set used in the MindCube ablations.
These files combine training questions from VSI-style tasks~\citep{yang2024vsi},
SpatialLadder~\citep{li2025spatialladder}, SPAR~\citep{zhang2025spar},
ViewSpatial~\citep{li2025viewspatial}, Spatial-MM~\citep{shiri2024spatialmm}, MMSI~\citep{yang2025mmsi},
and CV-Bench-style distance tasks~\citep{tong2024cambrian}.
The RGB images come primarily from ScanNet~\citep{dai2017scannet},
COCO~\citep{lin2014coco}, ScanNet++~\citep{yeshwanth2023scannetpp},
ARKitScenes~\citep{baruch2021arkitscenes},
Matterport3D~\citep{chang2017matterport3d}, and
nuScenes~\citep{caesar2020nuscenes}.
We exclude official benchmark test questions and answers from Stage~2
training. Sources with published training partitions are drawn from
those partitions; separately generated questions use the source training
pools. Stage~1 and Stage~2 training scenes do not overlap with
evaluation-benchmark scenes.
\cref{fig:training-data-pie} shows example inputs and the QA share of all
$14$ RGB sources.

\subsubsection{Geometry Sources}
\label{sec:appendix-geometry-inputs}
\paragraph{Predicted geometry.}
For predicted geometry, we use the Depth Anything 3~\citep{lin2025da3}
\texttt{DA3NESTED-GIANT-LARGE-1.1} checkpoint to estimate camera poses and
depth maps from RGB alone; main models use it for both training and
evaluation. \cref{tab:da3-geometry-error} reports the
errors of these predictions on training and evaluation images whose RGB
sources provide metric ground truth.
\begin{table}[!t]
\caption{Depth Anything 3~\citep{lin2025da3} geometry errors on training and evaluation images with ground truth. Rows group images by RGB source; A and B each sample up to 200 questions per
source and metric from Stage~2 training and evaluation, respectively. Depth
errors (AbsRel: mean absolute relative error; $\delta_1$: share of pixels
within a factor of 1.25 of ground truth) use valid ground-truth pixels without scale alignment. Pose errors are
relative to the earliest image in each question with valid ground-truth pose;
only subsequent valid images are scored: rotation error in degrees and
camera-position error in meters, without scale alignment. SUN RGB-D~\citep{song2015sunrgbd} images are
single views without pose (--). Matterport3D~\citep{chang2017matterport3d}, nuScenes~\citep{caesar2020nuscenes} and Omni3D~\citep{brazil2023omni3d} images are
omitted because they lack pixel-aligned metric ground truth. Errors are low on indoor scans and highest on SUN RGB-D.}
\label{tab:da3-geometry-error}
\centering\small
\setlength{\tabcolsep}{4pt}
\begin{tabular*}{\linewidth}{@{\extracolsep{\fill}}llcccc@{}}
\toprule
RGB source & Benchmarks & AbsRel (\%) $\downarrow$ & $\delta_1$ (\%) $\uparrow$ & Rot.\ ($^\circ$) $\downarrow$ & Trans.\ (m) $\downarrow$ \\
\tablegrouprule
\rowcolor{TableGroupFill}[0pt][0pt]
\multicolumn{6}{@{}l@{}}{\strut\tablegroup{A. Training images}} \\
\tablegroupgap
ScanNet & & 3.3 & 98.7 & 2.9 & 0.13 \\
\arrayrulecolor{black!30}\specialrule{0.35pt}{1.5pt}{1.5pt}\arrayrulecolor{black}
ScanNet++ & & 4.6 & 97.4 & 4.3 & 0.27 \\
\arrayrulecolor{black!30}\specialrule{0.35pt}{1.5pt}{1.5pt}\arrayrulecolor{black}
ARKitScenes & & 5.7 & 96.3 & 2.7 & 0.15 \\
\arrayrulecolor{black!30}\specialrule{0.35pt}{1.5pt}{1.5pt}\arrayrulecolor{black}
SUN RGB-D & & 26.1 & 89.3 & -- & -- \\
\tablegrouprule
\rowcolor{TableGroupFill}[0pt][0pt]
\multicolumn{6}{@{}l@{}}{\strut\tablegroup{B. Evaluation images}} \\
\tablegroupgap
ScanNet & \shortstack[l]{VSI, SPBench, ViewSpatial,\\SPAR, MMSI} & 5.9 & 97.4 & 2.8 & 0.11 \\
\arrayrulecolor{black!30}\specialrule{0.35pt}{1.5pt}{1.5pt}\arrayrulecolor{black}
ScanNet++ & VSI, SPAR & 8.8 & 92.3 & 2.7 & 0.18 \\
\arrayrulecolor{black!30}\specialrule{0.35pt}{1.5pt}{1.5pt}\arrayrulecolor{black}
ARKitScenes & VSI & 8.0 & 94.1 & 3.0 & 0.18 \\
\bottomrule
\end{tabular*}
\end{table}

Each cell reports the mean over questions, after averaging over valid images
or image pairs within a question. Depth and pose were sampled independently.
In the evaluation block, the ScanNet pose and ScanNet++ pose values use
175 and 195 valid questions, respectively, after excluding questions with
invalid native poses. The ScanNet depth sample is restricted to scenes with
locally available sensor data. ARKitScenes training depth is scored only on
the pixels covered by official depth after RGB--depth camera registration
(16\% of the RGB grid on average).

\paragraph{Annotated geometry.}
For ablations, we derive annotated geometry from the grounded
cognitive maps provided by MindCube~\citep{yin2025mindcube}. We extract only
each view's two-dimensional grid position and discrete facing direction,
without using object annotations, reasoning chains, or QA answers to construct
the geometric inputs. Views are matched to the ordered input images, and
their positions and orientations are expressed relative to the first view.
One grid cell defines one input unit; relative height, pitch, and roll are
set to zero. Intrinsics follow a fixed normalized convention
($f_x/W=f_y/H=1$, $c_x/W=c_y/H=0.5$), rather than camera calibration.
These inputs keep geometry-estimation errors out of the ablations; benchmark
comparisons use DA3 geometry.

\subsection{Model and Training Details}
\label{sec:appendix-model-training}
This section details the model and its training.
\Cref{sec:appendix-reader-architecture} describes the Geometric Reader
(\cref{sec:method-encoding}); \cref{sec:appendix-gate-questions} describes the
geometry injector, its alternatives in \cref{tab:read-then-inject-ablation} and
the two-pass computation (\cref{sec:method-injection}); and
\cref{sec:appendix-training-settings} gives the training settings of the main
models and the MindCube ablations.

\subsubsection{Geometric Reader Architecture}
\label{sec:appendix-reader-architecture}

\begin{figure}[!t]
  \centering
  \begin{minipage}[t]{0.57\linewidth}
    \vspace{0pt}
    \includegraphics[width=\linewidth]{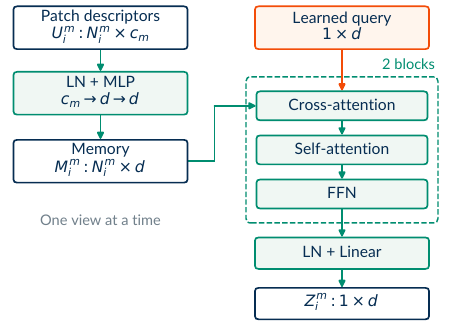}
  \end{minipage}\hfill
  \begin{minipage}[t]{0.40\linewidth}
    \vspace{6pt}
    \small
    \textbf{Reader configuration}\par\smallskip
    \begin{tabular*}{\linewidth}{@{}l@{\extracolsep{\fill}}r@{}}
      \toprule
      Hidden width $d$ & 2048 \\
      Blocks & 2 \\
      Attention heads & 8 \\
      FFN width & 8192 \\
      Activation & GELU \\
      Dropout & 0 \\
      Queries per view & 1 \\
      \bottomrule
    \end{tabular*}
    \par\medskip
    $N_i^m$: patch count in view $i$.\par
    $c_m$: descriptor dimension.\par
    LN: LayerNorm.
  \end{minipage}
  \caption{Architecture of a Geometric Reader. A descriptor MLP builds memory, read by a learned query through two attention blocks. Each signal's Reader shares weights across views, without cross-view attention.
  Outputs feed the Stage~2 geometry injector, one geometric token per view.}
  \label{fig:reader-architecture}
\end{figure}

\cref{fig:reader-architecture} summarizes the per-view Reader and its
Qwen2.5-VL-3B~\citep{bai2025qwen} configuration.

\paragraph{Per-view encoding.}
For view $i$ and geometric signal $m$, let
$U_i^m\in\mathbb{R}^{N_i^m\times c_m}$ contain the patch descriptors derived from the raw signal $G^m$.
The descriptor width $c_m$ is 20 for camera and 5 for depth.
The Reader forms memory
$M_i^m=\mathrm{MLP}_m(\mathrm{LN}(U_i^m))$,
where $\mathrm{MLP}_m$ is the two-layer descriptor MLP,
its linear maps act independently on the last dimension, and
$M_i^m\in\mathbb{R}^{N_i^m\times d}$.
A learned $1\times d$ query, shared across views, is updated by two
cross-attention--self-attention--feed-forward blocks.
Cross-attention uses normalized queries and memory for its query and key
projections, and the unnormalized memory for its value projection.
Self-attention and the feed-forward network use pre-LayerNorm;
all three sublayers have residual connections. Self-attention operates on
the query set within a view, not across views.
Final LayerNorm and a linear $d\to d$ projection give $Z_i^m$.
The Qwen2.5-VL-3B configuration uses $d=2048$, eight attention heads,
an $8192$-wide GELU feed-forward layer, and zero dropout.
The descriptor MLP has widths $c_m\to2048\to2048$.
Each signal's Reader uses the same block structure with independent weights.
During Stage~1, a learnable scalar output gate multiplies the Reader outputs
before they enter the frozen VLM.

\paragraph{Geometry descriptors.}
\cref{tab:camera-descriptors,tab:depth-descriptors}
list the 20-dimensional pose-derived and five-dimensional depth-derived
inputs, respectively. View $i=1$ is the reference view.
Displacements retain the units of their geometry source; annotated grid
units are not interpreted as meters.
\begin{table}[!t]
\centering
\small
\setlength{\tabcolsep}{4pt}
\caption{Camera descriptor components ($20$ dimensions per geometry patch). View~1 defines the reference frame; intrinsics are normalized by the processed image dimensions.}
\label{tab:camera-descriptors}
\begin{tabular*}{\linewidth}{@{\extracolsep{\fill}}p{0.27\linewidth}cp{0.62\linewidth}@{}}
\toprule
Component & Dim. & Meaning / encoding \\
\midrule
Image position & $2$ & Normalized patch-center coordinates $(u,v)$. \\
World ray & $3$ & Unit viewing ray in the source world frame. \\
Reference ray & $3$ & The same ray expressed in the reference-view frame. \\
Planar displacement & $2$ & Camera-center displacement $(\Delta x,\Delta z)$ relative to the reference view, in source units. \\
Viewing direction & $3$ & Camera forward direction in the reference-view frame. \\
Intrinsics & $4$ & $(f_x/W,f_y/H,c_x/W,c_y/H)$; $W,H$ are the processed image width and height. \\
View index & $1$ & $(i-1)/31$, using a fixed maximum of 32 views. \\
Reference flag & $1$ & 1 for view 1; 0 for other views. \\
Temporal position & $1$ & Normalized temporal patch center; $0.5$ for a still image. \\
\bottomrule
\end{tabular*}

\bigskip
\caption{Depth descriptor components ($5$ dimensions per geometry patch). Depth is measured in meters; the center value is log-transformed, and valid fraction records patch coverage.}
\label{tab:depth-descriptors}
\begin{tabular*}{\linewidth}{@{\extracolsep{\fill}}p{0.27\linewidth}cp{0.62\linewidth}@{}}
\toprule
Component & Dim. & Meaning / encoding \\
\midrule
Image position & $2$ & Normalized patch-center coordinates $(u,v)$. \\
View index & $1$ & $(i-1)/31$, using a fixed maximum of 32 views. \\
Center depth & $1$ & $\log(1+z)$ for valid center depth $z$; 0 otherwise. \\
Valid fraction & $1$ & Fraction of pixels with depth greater than $0.01$ m within the patch. \\
\bottomrule
\end{tabular*}
\end{table}

Depth is resized by nearest-neighbor interpolation to the processed image
grid; values above $0.01$ meters are valid, and invalid center depth is
encoded as zero. For this backbone, a geometry patch covers a $2\times2$
group of 14-pixel visual patches. Camera and depth descriptors share the
spatial grid and ordering of the merged RGB tokens, so $N_i^m$ equals the number of merged
RGB tokens of view~$i$.

\paragraph{View isolation.}
Variable patch counts are padded to the largest count within an example,
with padded keys excluded from cross-attention. The implementation treats
views as separate batch items inside the Reader, yielding
$[V,1,d]$ before flattening to $[1,V,d]$ in view order.
Reference-relative descriptors still encode cross-view relations, although
the aggregation never attends across views.

\paragraph{Output interface.}
For four views, the Reader outputs four embeddings
$Z_1^m,\ldots,Z_4^m\in\mathbb{R}^{1\times2048}$, one per view; in general,
each available signal supplies $V$ embeddings, independently of the number of
image patches.
Stage~2 uses these raw outputs, taken before the Stage~1 output gate, and
the injection designs apply their own scaling or adapter and gate
(\cref{sec:appendix-gate-questions}).

\subsubsection{Injection Design and Computation}
\label{sec:appendix-gate-questions}
\paragraph{Question-conditioned injector.}
\georeader{} follows \cref{sec:method-injection}: its adapter $P_m$ is a
tokenwise $2048\!\times\!2048$ affine map with zero-initialized weights and
bias, and its routing head $R_m$ applies LayerNorm, a $2048\to256$ linear map,
GELU, and a $256\to1$ linear map before the sigmoid. The final map is
zero-initialized, giving an initial gate of $0.5$ for available geometry.

\paragraph{Alternative designs (\cref{tab:read-then-inject-ablation}).}
\cref{tab:read-then-inject-ablation} replaces the whole injector with two
simpler designs, using annotated geometry and the same frozen pair-supervised
Reader. For the geometric signal evaluated here, let
$Z^m\in\mathbb{R}^{V\times d}$ denote the Reader's raw output, with one token
per view and $d=2048$. Direct injection inserts these outputs unchanged. Fixed scaling multiplies them by a single trainable
scalar, shared across examples and views, initialized to 1 and with no sigmoid
or normalization. All three designs bypass the Reader's Stage~1 output gate
(\cref{sec:appendix-reader-architecture}). On all 1,050 MindCube-Tiny
questions, direct injection, fixed scaling and the injector answer 825, 719
and 937 correctly.

\paragraph{Sequence and computation.}
For a four-view example with camera pose and depth, write $X_i$ for the RGB
token block of view~$i$, $Q_1,\ldots,Q_{K_q}$ for the $K_q$ question tokens, and
$\widetilde Z_i^{\mathrm{cam}}$ and $\widetilde Z_i^{\mathrm{dep}}$ for the
camera and depth geometric tokens of view~$i$. Omitting chat-template
delimiters, \georeader{} uses the two sequences
\begin{align*}
\text{Read:}\quad &[X_1][X_2][X_3][X_4][Q_1]\cdots[Q_{K_q}],\\
\text{Answer:}\quad &[X_1][X_2][X_3][X_4][Q_1]\cdots[Q_{K_q}]
[\widetilde Z_1^{\mathrm{cam}}]\cdots[\widetilde Z_4^{\mathrm{cam}}]
[\widetilde Z_1^{\mathrm{dep}}]\cdots[\widetilde Z_4^{\mathrm{dep}}].
\end{align*}
For $V$ views and $|\mathcal{M}|$ signals, these sequences contain
$\sum_i N_{\mathrm{rgb},i}+K_q$ and $\sum_i N_{\mathrm{rgb},i}+K_q+|\mathcal{M}|V$
embeddings, respectively, where
$N_{\mathrm{rgb},i}$ is the number of merged RGB tokens of view~$i$, excluding
template delimiters; each embedding has width 2048. Both training passes
use all 36 layers of the Qwen2.5-VL-3B language model with causal attention. The read pass
contains no geometry or answer tokens and extracts the detached layer-36
hidden state at the last valid question token as $h_q$. That state can see
the preceding RGB and question tokens. The answer pass inserts the geometric
embeddings before layer~1; geometric positions can attend to the preceding
RGB, question and geometric positions, while question positions cannot
attend to the subsequent geometry. Answer tokens follow the prompt and
can attend to all preceding inputs. Teacher forcing appends the answer
sequence and applies QA loss at answer prediction positions; at inference, each next token is predicted from the last position's final-layer hidden state.
Direct injection and fixed scaling run a single pass, whereas at inference
\georeader{} reuses the read-pass KV cache (\cref{sec:appendix-inference-cost}).

\subsubsection{Training Settings}
\label{sec:appendix-training-settings}
\paragraph{Shared settings.}
All runs use AdamW in bf16. Stage~2 fine-tunes the language model and the
geometry injector with learning rate $2\times10^{-5}$, 100 warm-up steps,
linear decay, weight decay~$0.01$ and image-area bounds of 3,136--150,528
pixels, and freezes the Readers and the visual encoder.

\paragraph{Stage~1 Readers.}
For every backbone, the Geometric Readers are trained in the same way: with
the VLM frozen, for 10 epochs on four GPUs, with per-GPU batch size~1,
gradient accumulation~1 and learning rate $10^{-4}$, on the Stage~1 data of
\cref{sec:appendix-stage1-data}.

\paragraph{Main models.}
The \georeader{} models in \cref{tab:generalization-results}, built on
3B/7B Qwen2.5-VL~\citep{bai2025qwen}, 4B/8B Qwen3-VL~\citep{bai2025qwen3vl}
and 4B/9B Qwen3.5~\citep{qwen2026qwen35}, are trained on the approximately
135K-example QA collection (\cref{sec:appendix-training-mixture}) for 2 epochs
on four GPUs, with per-GPU batch size~1 and gradient accumulation~2
(effective batch size~8; other settings as above).

\paragraph{MindCube ablations.}
Our QA-trained ablations in \cref{tab:alignment-ablation,tab:text-cot-comparison,tab:stage-results,tab:read-then-inject-ablation,tab:token-use-ablation}
start from Qwen2.5-VL-3B~\citep{bai2025qwen} and use the same 10K
MindCube~\citep{yin2025mindcube} QA examples for 10 epochs (100,000 optimizer
steps), each on one GPU with batch size~1 and no gradient accumulation.
Variants differ only in the compared component; e.g.,
fixed scaling in \cref{tab:read-then-inject-ablation} trains a
shared scalar in place of the geometry injector.

\subsection{Evaluation Protocols and Result Sources}
\label{sec:appendix-table1-sources}
\label{sec:appendix-eval-protocol}
This section describes how the reported results are obtained.
\Cref{sec:appendix-benchmark-metrics} defines the benchmarks and their
scoring, \cref{sec:appendix-local-eval,sec:appendix-api-eval} describe how we
evaluate local and API models, and \cref{sec:appendix-baseline-sources} gives
the sources of literature results in
\cref{tab:mindcube-ego3d-results,tab:generalization-results}.

\subsubsection{Benchmark Inputs and Metrics}
\label{sec:appendix-benchmark-metrics}
\paragraph{MindCube-Tiny.}
We follow the MindCube-Tiny benchmark~\citep{yin2025mindcube}.
The rotation, among, and around subsets contain 200, 600, and 250 questions,
respectively. For our evaluations, subtask scores and overall accuracy are
computed from the same 1,050 per-question predictions of each model; overall accuracy
is weighted by the three subset sizes.

\paragraph{Six benchmarks.}
Our evaluations of base models and \georeader{} use the full test sets:
5,130 VSI-Bench~\citep{yang2024vsi}, 1,328 SPBench~\citep{li2025spatialladder},
2,638 CV-Bench~\citep{tong2024cambrian}, 7,211 SPAR-Bench~\citep{zhang2025spar},
5,712 ViewSpatial~\citep{li2025viewspatial} and 1,000
MMSI-Bench~\citep{yang2025mmsi} questions. For VSI-Bench, 32 uniformly sampled
frames per video are supplied as ordered images. Each benchmark uses its
official scoring (SPBench: equal-weight mean of SI and MV), and Avg.\ in \cref{tab:generalization-results} averages the six
unrounded scores.

\subsubsection{Local-Model Evaluation and Scoring}
\label{sec:appendix-local-eval}
\paragraph{Generation.}
We run the Qwen base models (Qwen2.5-VL-3B/7B~\citep{bai2025qwen},
Qwen3-VL-2B/4B/8B~\citep{bai2025qwen3vl} and
Qwen3.5-0.8B/2B/4B/9B~\citep{qwen2026qwen35}) and \georeader{} locally with
native preprocessing, greedy decoding with at most 1,024 new tokens, and
image-area bounds of 3,136--150,528 pixels. All Qwen3.5 models run in non-thinking mode.

\paragraph{Answer extraction.}
For multiple-choice questions, we take the option letter inside the answer
tag, then one following ``Answer:'', and then a leading option letter; for
numerical questions, we take the first number in the response. Unextractable
answers count as incorrect on all benchmarks.

\subsubsection{API-Model Evaluation}
\label{sec:appendix-api-eval}
The two API models we evaluate, DeepSeek-Flash and MiMo-V2.6-Flash, use their
own image processing and are evaluated unadapted in September 2026 on all
1,050 MindCube-Tiny questions, with original images in dataset order, the official
raw-QA instruction, and the answer extraction in \cref{sec:appendix-local-eval}.
Their 65,536-token limits include reasoning and final output.

\paragraph{\includegraphics[height=7pt]{figures/model_logos/deepseek-color.pdf}\,DeepSeek-V4.1-Flash~\citep{deepseek2026v41flash}.}
It appears as DeepSeek-Flash in \cref{tab:mindcube-ego3d-results}. We
request \texttt{deepseek-flash} with high reasoning effort and
\texttt{original} image detail. Thinking mode ignores temperature. Four
responses hit the limit; five unextractable answers count as incorrect.

\paragraph{\includegraphics[height=7pt]{figures/model_logos/xiaomimimo.pdf}\,MiMo-V2.6-Flash~\citep{xiaomi2026mimov26flash}.}
We request \texttt{mimo-v2.6-flash} with deep thinking on; it has no
image-detail option. No response reaches the limit; 26
unextractable answers count as incorrect.

\paragraph{Model sizes.}
Sizes in parentheses in \cref{tab:mindcube-ego3d-results} are total parameters
reported by the provider. \includegraphics[height=7pt]{figures/model_logos/doubao-color.pdf}\,Doubao-Seed-1.6 has 230B
total and 23B active~\citep{bytedance2025seed16}, DeepSeek-V4.1-Flash
552B total, and MiMo-V2.6-Flash 309B total and 15B active parameters~\citep{xiaomi2026mimov26flash}.

\subsubsection{Sources of Baseline Results}
\label{sec:appendix-baseline-sources}
\paragraph{\cref{tab:mindcube-ego3d-results}.}
We evaluated the Qwen base models~\citep{bai2025qwen,bai2025qwen3vl,qwen2026qwen35},
\includegraphics[height=7pt]{figures/model_logos/deepseek-color.pdf}\,DeepSeek-Flash~\citep{deepseek2026v41flash} in
thinking mode, \includegraphics[height=7pt]{figures/model_logos/xiaomimimo.pdf}\,MiMo-V2.6-Flash~\citep{xiaomi2026mimov26flash}
and all \georeader{} rows. The SenseNova-SI~\citep{sensenova2025} results for
Qwen2.5-VL-3B and 7B come from the official
\href{https://huggingface.co/sensenova/SenseNova-SI-1.1-Qwen2.5-VL-3B}{\linkfont 3B}
and
\href{https://huggingface.co/sensenova/SenseNova-SI-1.1-Qwen2.5-VL-7B}{\linkfont 7B}
model cards (version~1.1), and the 3DThinker~\citep{chen2026think3d} results
with Qwen3-VL-2B/4B/8B use the two-stage (S1+S2) rows in Tab.~1 of its CVPR 2026
supplementary material. The other rows use the results reported in the cited
papers.

\paragraph{\cref{tab:generalization-results}.}
We evaluated the Qwen base models marked $^{\dagger}$ and all \georeader{}
rows. For SpatialLadder~\citep{li2025spatialladder}, the VSI-Bench, CV-Bench,
SPAR-Bench, ViewSpatial and MMSI-Bench scores come from Tabs.~1--2 of its
ICLR 2026 paper; that paper reports SPBench-SI and SPBench-MV separately, so
its combined SPBench score comes from Tab.~2 of 3DThinker~\citep{chen2026think3d}.
For ViLaSR~\citep{wu2025vilasr}, the VSI-Bench, SPAR-Bench and MMSI-Bench
scores come from Tab.~1 of its NeurIPS 2025 paper and the other three from
3DThinker Tab.~2. The Spatial-MLLM~\citep{wu2025spatialmllm},
SpaceR~\citep{ouyang2025spacer}, Video-R1~\citep{feng2025videor1} and
3DThinker rows come entirely from 3DThinker Tab.~2. The
SenseNova-SI-8B~\citep{sensenova2025} row combines VSI-Bench, MMSI-Bench and
ViewSpatial from the official
\href{https://huggingface.co/sensenova/SenseNova-SI-1.1-Qwen3-VL-8B}{\linkfont Qwen3-VL-8B version~1.1 model card}
with CV-Bench, SPAR-Bench and SPBench-SI/MV from GAMSI~\citep{zheng2026gamsi}
Tab.~1 (release unspecified); its SPBench-SI/MV scores
($60.4/71.1$) and six-benchmark average are combined as in
\cref{sec:appendix-benchmark-metrics}. Literature results keep their source evaluation settings.

\subsection{Additional Experiments}
\label{sec:appendix-additional-comparisons}
These experiments follow the order of
\crefrange{sec:spatial-performance}{sec:interface-tradeoffs}.
\Cref{sec:appendix-rgb-only-comparison,sec:appendix-backbone-results} extend
the main results with a subtask comparison against vanilla SFT and with
Qwen3.5 on MindCube-Tiny; \cref{sec:appendix-geometry-sources} compares
predicted and annotated geometry;
\cref{sec:appendix-pretraining-controls,sec:appendix-adapted-fusion} detail the
Reader-pretraining and alternative-interface comparisons;
\cref{sec:appendix-content-protocol} describes the geometry interventions; and
\cref{sec:appendix-text-interface-protocol} covers the text-geometry recipes
and inference cost.

\begin{table}[!t]
\caption{Subtask-level accuracy (\%) of vanilla SFT and \georeader{}. Both models use Qwen2.5-VL-3B~\citep{bai2025qwen} with the same QA data and training settings.
A and B split all 59 native subtasks of the six benchmarks by whether the
question requires metric geometry (number of subtasks in parentheses),
with accuracies weighted by question count. C lists every subtask in A whose
gain is at least $2$ points. Images: input RGB images or sampled video frames (ranges indicate mixed input sizes); Questions: number of questions.
$\Delta$ is computed before rounding. Gains concentrate on metric 3D subtasks, and the other subtasks are unchanged.}
\label{tab:subtask-gains}
\centering\small
\setlength{\tabcolsep}{3pt}
\begin{tabular*}{\linewidth}{@{\extracolsep{\fill}}llcrccc@{}}
\toprule
\multicolumn{2}{@{}l}{Question type} & Images & Questions & Vanilla & \ours{} & $\Delta$ \\
\tablegrouprule
\rowcolor{TableGroupFill}[0pt][0pt]
\multicolumn{7}{@{}l@{}}{\rule[-0.8ex]{0pt}{3.0ex}\tablegroup{A. Questions requiring metric geometry}} \\
\tablegroupgap
\multicolumn{2}{@{}l}{Object and room size (4)} & $1\text{--}32$ & $1{,}862$ & $65.2$ & $70.0$ & $+4.8$ \\
\multicolumn{2}{@{}l}{Camera motion (4)} & $2$ & $967$ & $39.3$ & $43.7$ & $+4.4$ \\
\multicolumn{2}{@{}l}{Distance and depth (19)} & $1\text{--}32$ & $6{,}613$ & $56.8$ & $59.0$ & $+2.2$ \\
\cmidrule{1-7}
\multicolumn{2}{@{}l}{Metric-geometry questions (27)} & $1\text{--}32$ & $9{,}442$ & $56.7$ & $59.6$ & $\mathbf{+2.9}$ \\
\tablegrouprule
\rowcolor{TableGroupFill}[0pt][0pt]
\multicolumn{7}{@{}l@{}}{\rule[-0.8ex]{0pt}{3.0ex}\tablegroup{B. Other questions}} \\
\tablegroupgap
\multicolumn{2}{@{}l}{Direction, relation and other (27)} & $1\text{--}32$ & $10{,}886$ & $65.5$ & $65.6$ & $+0.1$ \\
\multicolumn{2}{@{}l}{2D, counting and order (5)} & $1\text{--}32$ & $2{,}691$ & $74.3$ & $75.1$ & $+0.8$ \\
\cmidrule{1-7}
\multicolumn{2}{@{}l}{All other questions (32)} & $1\text{--}32$ & $13{,}577$ & $67.3$ & $67.5$ & $\mathbf{+0.3}$ \\
\tablegrouprule
\rowcolor{TableGroupFill}[0pt][0pt]
\multicolumn{7}{@{}l@{}}{\rule[-0.8ex]{0pt}{3.0ex}\tablegroup{C. Metric-geometry subtasks with gains of at least 2 points}} \\
\tablegroupgap
Benchmark & Subtask & Images & Questions & Vanilla & \ours{} & $\Delta$ \\
\midrule
\multirow{4}{*}{VSI-Bench} & Object size & $32$ & $953$ & $64.5$ & $71.9$ & $+7.4$ \\
 & Room size & $32$ & $288$ & $55.4$ & $62.0$ & $+6.7$ \\
 & Absolute distance & $32$ & $834$ & $38.4$ & $41.8$ & $+3.4$ \\
 & Relative distance & $32$ & $710$ & $47.9$ & $50.6$ & $+2.7$ \\
\arrayrulecolor{black!30}\specialrule{0.35pt}{1.5pt}{1.5pt}\arrayrulecolor{black}
\multirow{3}{*}{SPBench} & Relative distance & $1$ & $91$ & $69.2$ & $83.5$ & $+14.3$ \\
 & Object size & $8$ & $158$ & $69.3$ & $72.9$ & $+3.6$ \\
 & Absolute distance & $8$ & $30$ & $38.0$ & $40.3$ & $+2.3$ \\
\arrayrulecolor{black!30}\specialrule{0.35pt}{1.5pt}{1.5pt}\arrayrulecolor{black}
MMSI & Camera motion & $2$ & $74$ & $43.2$ & $56.8$ & $+13.5$ \\
\arrayrulecolor{black!30}\specialrule{0.35pt}{1.5pt}{1.5pt}\arrayrulecolor{black}
\multirow{7}{*}{SPAR} & Object--object distance & $1$ & $363$ & $48.1$ & $54.2$ & $+6.0$ \\
 & Distance inference & $1$ & $340$ & $73.2$ & $77.6$ & $+4.4$ \\
 & Object--camera distance & $1$ & $393$ & $64.6$ & $68.2$ & $+3.7$ \\
 & Object--camera depth & $1$ & $360$ & $69.6$ & $71.9$ & $+2.3$ \\
 & Camera-motion inference & $2$ & $400$ & $49.8$ & $57.5$ & $+7.8$ \\
 & Distance inference & $3$ & $336$ & $63.4$ & $69.3$ & $+6.0$ \\
 & Object--object distance & $3$ & $297$ & $38.3$ & $41.0$ & $+2.7$ \\
\arrayrulecolor{black!30}\specialrule{0.35pt}{1.5pt}{1.5pt}\arrayrulecolor{black}
CV-Bench & Depth & $1$ & $600$ & $86.5$ & $88.8$ & $+2.3$ \\
\bottomrule
\end{tabular*}
\end{table}

\subsubsection{Geometry versus Vanilla SFT on Six Benchmarks}
\label{sec:appendix-rgb-only-comparison}
We compare \georeader{}-3B from \cref{tab:generalization-results} with
vanilla SFT, the same QA fine-tuning without geometric inputs, to separate the
effect of geometry from that of QA fine-tuning. Both use
Qwen2.5-VL-3B~\citep{bai2025qwen} with the same QA data, training settings and
evaluation protocol; \georeader{} also reads DA3-predicted geometry through its frozen
Readers.

Each benchmark mixes questions that depend on 3D geometry with questions that
do not, so we split the 59 native subtasks of the six benchmarks into those
requiring metric geometry (object and room size, camera motion, and distance
and depth) and the others (direction, relation and other; 2D, counting and
order), grouped by the quantity asked about. Camera motion
covers questions about the camera's own motion or viewpoint change (MMSI
camera motion and camera--camera position, SPAR camera-motion and view-change
inference); the size and distance-and-depth types cover subtasks named by
those quantities; 2D, counting and order covers 2D relations, counting and
appearance order; the remaining subtasks form direction, relation and other.
Question-type scores pool subtasks across benchmarks and weight each subtask
by its number of questions, whereas benchmark totals follow each benchmark's
own averaging (e.g., SPBench averages its SI and MV splits equally), so
question-type scores are not directly comparable with the totals in
\cref{tab:generalization-results}.

On the 27 subtasks requiring metric geometry, geometric inputs raise accuracy
from $56.7$ to $59.6$, and every question type improves: object and room size
($+4.8$), camera motion ($+4.4$) and distance and depth ($+2.2$)
(\cref{tab:question-type-gains,tab:subtask-gains}). Individual subtasks gain
much more, including $+14.3$ on SPBench-SI relative distance, $+13.5$ on MMSI
camera motion, $+7.8$ on SPAR camera-motion inference and $+7.4$ on VSI-Bench
object size; the subtasks in Panel~C of \cref{tab:subtask-gains} span 1, 2, 3,
8 and 32 input images or sampled frames per question. On the other 32
subtasks, accuracy is unchanged ($67.3$ versus $67.5$), and neither question
type decreases. Geometric inputs thus improve accuracy on geometry-dependent
questions without lowering it on the others.

\begin{table}[!t]
  \caption{MindCube-Tiny~\citep{yin2025mindcube} accuracy (\%) with Qwen3.5 backbones. Base models are evaluated by us, Video-ORA~\citep{li2026orarl} uses its published result, bold marks the best Overall, and \georeader{} uses Depth Anything 3~\citep{lin2025da3} geometry. \georeader{} improves over its Qwen3.5-4B~\citep{qwen2026qwen35} base model.}
  \label{tab:small-backbone-results}
  \centering
  \small
  \setlength{\tabcolsep}{3pt}
  \begin{tabular*}{\linewidth}{@{\extracolsep{\fill}}lcccc@{}}
    \toprule
    Model & Rot. & Among & Around & Overall $\uparrow$ \\
    \midrule
    Qwen3.5-0.8B~\citep{qwen2026qwen35} & $25.0$ & $31.8$ & $44.8$ & $33.6$ \\
    Qwen3.5-2B~\citep{qwen2026qwen35} & $40.5$ & $31.8$ & $51.2$ & $38.1$ \\
    \midrule
    Qwen3.5-4B~\citep{qwen2026qwen35} & $47.5$ & $43.2$ & $59.2$ & $47.8$ \\
    Video-ORA-4B~\citep{li2026orarl} & $37.0$ & $49.8$ & $54.8$ & $48.6$ \\
    \ours{} (Qwen3.5-4B) & $35.5$ & $93.5$ & $92.4$ & $\mathbf{82.2}$ \\
    \bottomrule
  \end{tabular*}
\end{table}

\subsubsection{MindCube-Tiny Results with Qwen3.5}
\label{sec:appendix-backbone-results}
\cref{tab:mindcube-ego3d-results} covers the Qwen2.5-VL and Qwen3-VL
backbones; \cref{tab:small-backbone-results} reports MindCube-Tiny results for
Qwen3.5-4B~\citep{qwen2026qwen35}, with Video-ORA-4B~\citep{li2026orarl} and
smaller Qwen3.5 base models (evaluated as in \cref{sec:appendix-eval-protocol})
for reference. \georeader{} raises Overall from $47.8$ for its base
model to $82.2$, above Video-ORA-4B ($48.6$), while the Qwen3.5 base models
reach $33.6$--$47.8$. The gains come from Among ($43.2$ to $93.5$) and Around
($59.2$ to $92.4$), whereas Rotation drops from $47.5$ to $35.5$; Qwen3.5-4B
is the only backbone where \georeader{} lowers Rotation
(\cref{tab:mindcube-ego3d-results}).

\begin{table}[!t]
\centering
\begin{minipage}[t]{0.48\linewidth}
\vspace{0pt}
\caption{MindCube-Tiny~\citep{yin2025mindcube} accuracy (\%) with DA3~\citep{lin2025da3} geometry across recipes; \cref{tab:text-cot-comparison} gives the annotated-geometry results. Pair-supervised \ours{} remains the best recipe.\strut}
\label{tab:geometry-source-comparison}
\end{minipage}\hfill
\begin{minipage}[t]{0.48\linewidth}
\vspace{0pt}
\caption{MindCube-Tiny~\citep{yin2025mindcube} accuracy (\%) of pair-supervised \ours{} with annotated or Depth Anything 3~\citep{lin2025da3} geometry. Bold marks the better source, which differs by backbone.\strut}
\label{tab:backbone-geometry-source-comparison}
\end{minipage}\par
\begin{minipage}[t]{0.48\linewidth}
\vspace{0pt}
\centering\small
\renewcommand{\arraystretch}{1.16} %
\begin{tabular*}{\linewidth}{@{\extracolsep{\fill}}lc@{}}
\toprule
Training and inference recipe & DA3 \\
\midrule
Joint description + QA SFT & $64.7$ \\
Generated Text-CoT & $66.7$ \\
Descriptions in prompt & $79.2$ \\
\ours{} (group) & $68.9$ \\
\ours{} (pair) & $\mathbf{81.1}$ \\
\bottomrule
\end{tabular*}
\end{minipage}\hfill
\begin{minipage}[t]{0.48\linewidth}
\vspace{0pt}
\centering\small
\begin{tabular*}{\linewidth}{@{\extracolsep{\fill}}lcc@{}}
\toprule
Backbone & Annotated & DA3 \\
\midrule
Qwen2.5-VL-3B & $\mathbf{89.2}$ & $81.1$ \\
Qwen2.5-VL-7B & $85.6$ & $\mathbf{93.2}$ \\
Qwen3-VL-2B   & $77.0$ & $\mathbf{86.5}$ \\
Qwen3-VL-4B   & $92.3$ & $\mathbf{94.3}$ \\
Qwen3-VL-8B   & $92.5$ & $\mathbf{93.4}$ \\
Qwen3.5-4B    & $\mathbf{83.4}$ & $82.2$ \\
\bottomrule
\end{tabular*}
\end{minipage}
\end{table}

\subsubsection{Predicted versus Annotated Geometry}
\label{sec:appendix-geometry-sources}
\label{sec:appendix-geometry-comparison}
Our ablations use annotated geometry to exclude estimation errors, whereas the
main models use DA3-predicted geometry. This comparison asks how switching to
predicted geometry changes accuracy and whether the ablation conclusions still
hold. Each configuration applies one source (\cref{sec:appendix-geometry-inputs})
to description targets, Reader training, QA training and evaluation.

With Qwen2.5-VL-3B and pair supervision, DA3 lowers accuracy from $89.2\%$ to
$81.1\%$, still above vanilla SFT ($68.6\%$;
\cref{tab:text-cot-comparison,tab:geometry-source-comparison}). Across the six
backbones in \cref{tab:backbone-geometry-source-comparison}, DA3 exceeds
annotated geometry on four (by $0.9$--$9.5$ points) and trails it on two (by
$1.2$--$8.1$ points). Neither source is better on all backbones, although DA3
is $1.8$ points higher on average.

With DA3 geometry (\cref{tab:geometry-source-comparison}), the recipes rank as
with annotated geometry (\cref{tab:text-cot-comparison}): pair supervision
$81.1\%$, descriptions in the prompt $79.2\%$, group supervision $68.9\%$,
generated Text-CoT $66.7\%$ and joint description + QA SFT $64.7\%$. The
ablation conclusions drawn with annotated geometry therefore also hold with
geometry predicted from RGB.

\subsubsection{Reader Pretraining}
\label{sec:appendix-pretraining-controls}

\paragraph{Stage results (\cref{tab:stage-results}).}
The Stage~1-only reference skips QA training: the frozen VLM receives the
Stage~1 geometric tokens, injected with a fixed scale of $0.25$. Vanilla SFT uses
the same 10K QA examples, and the full recipe uses pair-supervised annotated
geometry.

\paragraph{Feature alignment (\cref{tab:alignment-ablation}, Panel A).}
The feature-alignment row uses the released first-stage
3DThinker~\citep{chen2026think3d} checkpoint on Qwen2.5-VL-3B~\citep{bai2025qwen}
(\cref{sec:appendix-content-protocol}); its $62.7\%$ is computed from the
released per-question answers on the same 1,050 questions.

\begin{figure}[!t]
  \centering
  \includegraphics[width=\linewidth]{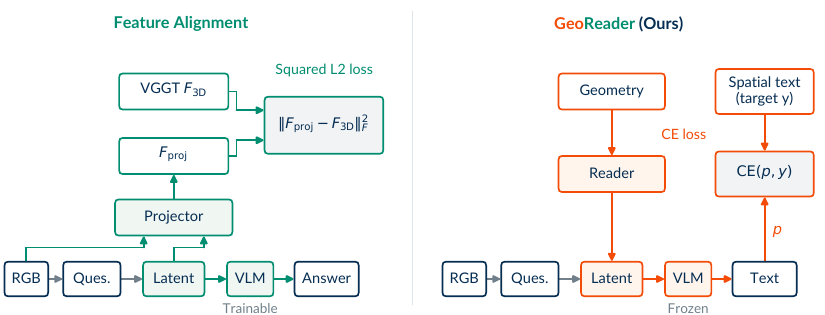}
  \caption{Geometry supervision in feature alignment and in \georeader{}. 3DThinker~\citep{chen2026think3d} aligns the latents of a trainable VLM with VGGT~\citep{wang2025vggt} features;
\georeader{} trains the Reader through a frozen VLM using geometry-description cross-entropy. Only geometry-supervision paths are shown; 3DThinker's
text loss is omitted. Ques. is the text prompt given to the VLM.}
  \label{fig:feature-alignment}
\end{figure}

\paragraph{Comparison with feature alignment.}
\label{sec:appendix-interface-comparison}
\cref{fig:feature-alignment} contrasts the geometry-supervision interfaces of
3DThinker~\citep{chen2026think3d} and \georeader{}. In 3DThinker, RGB features and
VLM-generated latents jointly enter a projector supervised by VGGT~\citep{wang2025vggt} features;
VGGT features are training targets, not additional inference inputs.
\georeader{} instead learns geometric inputs using geometry-description supervision through the frozen target VLM, then freezes the Reader for QA.
The figure contrasts these learning targets and input pathways side by side.

\subsubsection{Alternative Interfaces}
\label{sec:appendix-adapted-fusion}
\paragraph{Shared setting.}
The Panel~B alternatives in \cref{tab:alignment-ablation} follow the MindCube
ablation setting of \cref{sec:appendix-training-settings} with answer-only
supervision. The two adapted fusion rows use pre-extracted final-layer
VGGT~\citep{wang2025vggt} features from the input RGB images: one camera token
and 1,369 patch tokens per view, each 2,048-dimensional; the four register
tokens are excluded. The VGGT and Qwen visual encoders are frozen, while the
full language model and fusion module are trained with QA answer loss. Both
fusion modules use per-frame cross-attention with a 512-dimensional attention
space and eight heads. They modify the Qwen image embeddings before one LLM
pass, without adding geometric tokens to the input or using a Geometric Reader.

\paragraph{VLM-3R-style adaptation.}
Following the spatial--visual fusion idea of VLM-3R~\citep{fan2025vlm3r}, each
frame's Qwen image embeddings query that frame's VGGT camera and patch
features. A learned output projection maps the attended features back to the
Qwen hidden dimension, and its zero-initialized residual is added to the
corresponding image embeddings. We adopt only the fusion mechanism; the
original VLM-3R system uses CUT3R~\citep{wang2025cut3r} features.

\paragraph{SpaceMind-style adaptation.}
Inspired by the camera-guided fusion of SpaceMind~\citep{zhao2025spacemind},
we retain the same per-frame visual queries and VGGT features but use the
camera feature to adjust patch keys and values, weight patch values by learned
importance, and apply a camera-conditioned channel-wise gate to the fused
residual before adding it to the Qwen image embeddings. Its output projection
is zero-initialized, so the residual starts at zero.

\paragraph{Per-token additive fusion.}
This row uses the same annotated pose input as \ours{}: the 20-dimensional
descriptor of each image token passes through a zero-initialized linear map to
the Qwen hidden dimension and is added to its embedding. It adds no
geometric tokens and uses one LLM pass, whereas the Reader of \ours{} encodes the same
descriptors into one token per view.

\subsubsection{Geometry Interventions}
\label{sec:appendix-content-protocol}
\paragraph{\georeader{}.}
\cref{tab:token-use-ablation} uses the pair-supervised MindCube checkpoint
from \cref{tab:text-cot-comparison}, with identical weights, questions, images
and decoding settings across conditions. Correct geometry yields $937/1,050$
correct answers. Zero restores each geometric token to its slot embedding after
the Reader, adapter and gate have been evaluated; it keeps the two-pass
computation and yields $811/1,050$. Shuffled applies a within-example
permutation with no fixed points (seed 42) to the raw extrinsics and
intrinsics, leaving the RGB order unchanged; the original feature builder then
regenerates frame identifiers and the reference frame from the newly ordered
poses. This condition tests pose-to-view mismatch under the model's normal
feature construction, including its first-view reference convention, and
yields $663/1,050$. No condition has answer-parsing failures.

\paragraph{Feature alignment (Feat.\ align.).}
We use the released 3DThinker~\citep{chen2026think3d} checkpoint described in
\cref{sec:appendix-answer-attention}, which is the model after its first
training stage; the $75.2\%$ in \cref{tab:mindcube-ego3d-results} is the
published result after both stages, whose checkpoint is not released. Correct
uses the released per-question answers on all 1,050 questions ($658$ correct,
$62.7\%$). Zero replaces the input embedding at every latent-generation step
with a zero vector while keeping the released visible prefix, then decodes
greedily ($660$ correct, $62.9\%$). Shuffled replaces its latents with those
generated for another question, using a fixed one-to-one assignment in which
the two questions come from different scenes and have different correct
options ($646$ correct, $61.5\%$). Unlike \georeader{}'s within-question pose shuffle, it swaps latents across
questions. Invalid answers count as wrong in both interventions.

\subsubsection{Text-Geometry Recipes and Inference Cost}
\label{sec:appendix-text-interface-protocol}
\label{sec:appendix-inference-cost}
All six recipes in \cref{tab:text-cot-comparison} use the MindCube ablation
setting of \cref{sec:appendix-training-settings} with annotated geometry.

\paragraph{Training.}
Joint description + QA SFT adds 2,072 whole-group description examples per
epoch to the QA data. \georeader{} (group) and \georeader{} (pair) train their
Readers on 2,072 and 3,375 Stage~1 examples per epoch, respectively;
reference--target pairs follow \cref{sec:appendix-stage1-data}.

\paragraph{Inference.}
Generated Text-CoT receives the images and the original question and generates
descriptions before the answer. Descriptions in prompt and \georeader{}
additionally receive the corresponding geometry, as text and as geometric
tokens, respectively. Generation is limited to 128 tokens for direct answers
and to 512 tokens for Generated Text-CoT, including the answer.

\paragraph{Cost.}
\cref{tab:text-cot-comparison} reports generated tokens and latency for its six
recipes, measured on the same NVIDIA H200 with batch size~1. After five
excluded warm-up examples per model, we average over all 1,050 test questions.
Token counts include formatting, control and end-of-sequence tokens but
not input tokens. Timing includes image
loading, input processing, all VLM passes, generation and decoding; model
loading and offline geometry construction are excluded. For \georeader{},
latency includes both passes, with the answer pass reusing the read pass's
key--value cache (\cref{sec:method-injection}). The SFT baselines use standard
generation, and \georeader{} a custom greedy decoding loop.

\FloatBarrier
\subsection{Qualitative Analysis}
\label{sec:appendix-qualitative}
This section examines individual questions: answer-token attention of
\georeader{} and feature alignment (\cref{sec:appendix-answer-attention}),
additional correct examples (\cref{sec:appendix-predicted-examples}), and
failure cases (\cref{sec:appendix-failure-cases}).

\subsubsection{Answer-Token Attention}
\label{sec:appendix-answer-attention}
\begin{figure}[p]
\noindent\begin{minipage}{\linewidth}
\setlength{\fboxsep}{4pt}\setlength{\fboxrule}{1.6pt}
\noindent\begin{tikzpicture}
\node[draw=black!18,line width=1.6pt,rounded corners=6pt,inner sep=4pt,outer sep=0pt] {%
\begin{minipage}{\dimexpr\linewidth-2\fboxsep-\fboxrule\relax}
\fontencoding{T1}\fontfamily{lato-TLF}\mdseries\selectfont\raggedright
{\fontsize{9.5}{10.5}\selectfont\textbf{Among}}\par\vspace{2pt}
\noindent\begin{minipage}[t]{0.40\linewidth}\vspace{0pt}
\begin{minipage}[t]{0.240\linewidth}\centering \roundedimage{\includegraphics[width=\linewidth,height=1.15in,keepaspectratio]{figures/attention_cases/among/image_1.jpg}}\end{minipage}\hfill\begin{minipage}[t]{0.240\linewidth}\centering \roundedimage{\includegraphics[width=\linewidth,height=1.15in,keepaspectratio]{figures/attention_cases/among/image_2.jpg}}\end{minipage}\hfill\begin{minipage}[t]{0.240\linewidth}\centering \roundedimage{\includegraphics[width=\linewidth,height=1.15in,keepaspectratio]{figures/attention_cases/among/image_3.jpg}}\end{minipage}\hfill\begin{minipage}[t]{0.240\linewidth}\centering \roundedimage{\includegraphics[width=\linewidth,height=1.15in,keepaspectratio]{figures/attention_cases/among/image_4.jpg}}\end{minipage}
\end{minipage}\hfill\begin{minipage}[t]{0.58\linewidth}\vspace{0pt}
\fontsize{9}{10.2}\selectfont\raggedright\color{black!70}
Question. Based on these four images (image 1, 2, 3, and 4) showing the \figkeyword{black sneaker} from different viewpoints (front, left, back, and right), with each camera aligned with room walls and partially capturing the surroundings: From the viewpoint presented in image 2, what is to the right of the \figkeyword{black sneaker}?\par\vspace{2pt}
Options. A. TV B. Wooden dining table C. \figkeyword{Light purple sofa} D. Brown curtains and windows\par\vspace{2pt}
\correctanswer{\georeaderfig{}: C}\hspace{14pt}\correctanswer{3DThinker: C}
\end{minipage}
\par
\includegraphics[width=\linewidth]{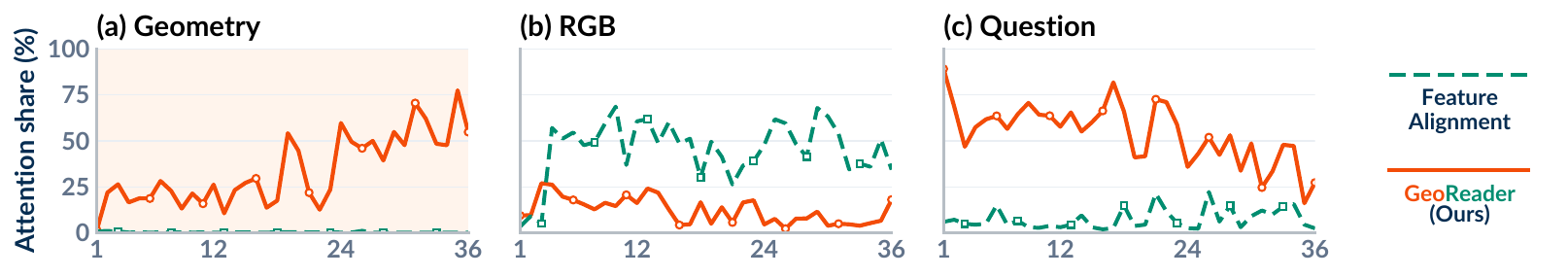}
\end{minipage}};
\end{tikzpicture}\par\vspace{3pt}

\noindent\begin{tikzpicture}
\node[draw=black!18,line width=1.6pt,rounded corners=6pt,inner sep=4pt,outer sep=0pt] {%
\begin{minipage}{\dimexpr\linewidth-2\fboxsep-\fboxrule\relax}
\fontencoding{T1}\fontfamily{lato-TLF}\mdseries\selectfont\raggedright
{\fontsize{9.5}{10.5}\selectfont\textbf{Around}}\par\vspace{2pt}
\noindent\begin{minipage}[t]{0.40\linewidth}\vspace{0pt}
\begin{minipage}[t]{0.320\linewidth}\centering \roundedimage{\includegraphics[width=\linewidth,height=1.15in,keepaspectratio]{figures/attention_cases/around/image_1.png}}\end{minipage}\hfill\begin{minipage}[t]{0.320\linewidth}\centering \roundedimage{\includegraphics[width=\linewidth,height=1.15in,keepaspectratio]{figures/attention_cases/around/image_2.png}}\end{minipage}\hfill\begin{minipage}[t]{0.320\linewidth}\centering \roundedimage{\includegraphics[width=\linewidth,height=1.15in,keepaspectratio]{figures/attention_cases/around/image_3.png}}\end{minipage}
\end{minipage}\hfill\begin{minipage}[t]{0.58\linewidth}\vspace{0pt}
\fontsize{9}{10.2}\selectfont\raggedright\color{black!70}
Question. Based on these three images (image 1, 2, and 3) showing the same scene from different viewpoints (back, left, and right): From the viewpoint presented in image 2, is there a \figkeyword{large blue recycling bin} behind the \figkeyword{small blue recycling bin}?\par\vspace{2pt}
Options. A. Yes B. No.\par\vspace{2pt}
\correctanswer{\georeaderfig{}: A}\hspace{14pt}\correctanswer{3DThinker: A}
\end{minipage}
\par
\includegraphics[width=\linewidth]{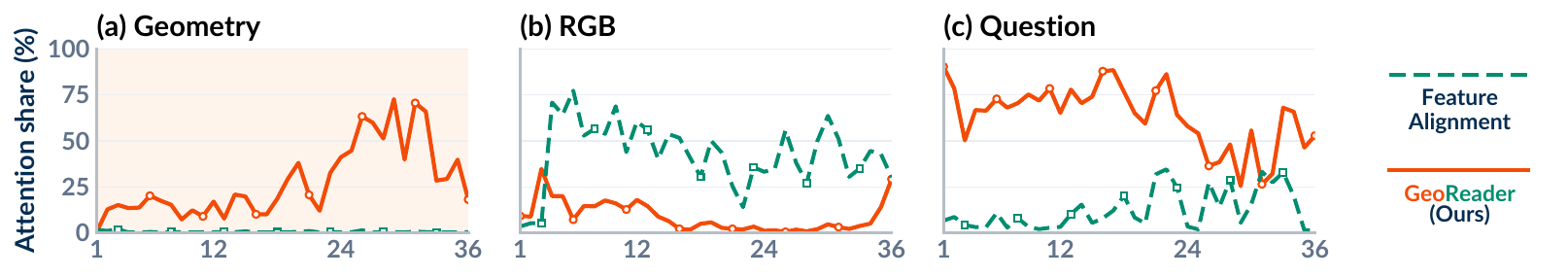}
\end{minipage}};
\end{tikzpicture}\par\vspace{3pt}

\noindent\begin{tikzpicture}
\node[draw=black!18,line width=1.6pt,rounded corners=6pt,inner sep=4pt,outer sep=0pt] {%
\begin{minipage}{\dimexpr\linewidth-2\fboxsep-\fboxrule\relax}
\fontencoding{T1}\fontfamily{lato-TLF}\mdseries\selectfont\raggedright
{\fontsize{9.5}{10.5}\selectfont\textbf{Rotation}}\par\vspace{2pt}
\noindent\begin{minipage}[t]{0.40\linewidth}\vspace{0pt}
\begin{minipage}[t]{0.320\linewidth}\centering \roundedimage{\includegraphics[width=\linewidth,height=1.15in,keepaspectratio]{figures/attention_cases/rotation/image_1.png}}\end{minipage}\hfill\begin{minipage}[t]{0.320\linewidth}\centering \roundedimage{\includegraphics[width=\linewidth,height=1.15in,keepaspectratio]{figures/attention_cases/rotation/image_2.png}}\end{minipage}\hfill\begin{minipage}[t]{0.320\linewidth}\centering \roundedimage{\includegraphics[width=\linewidth,height=1.15in,keepaspectratio]{figures/attention_cases/rotation/image_3.png}}\end{minipage}
\end{minipage}\hfill\begin{minipage}[t]{0.58\linewidth}\vspace{0pt}
\fontsize{9}{10.2}\selectfont\raggedright\color{black!70}
Question. These three images (image 1, 2, and 3) show the same scene from three different viewpoints. The image 2 was taken after turning the camera 90 degrees to the right (clockwise) from the position of image 1. For image 3, the camera was turned another 90 degrees right, so it's basically facing the opposite direction of image 1. Based on these three images: If I am standing at the same spot and facing the same direction as shown in image 3, then I turn 180 degrees around, what is to my right?\par\vspace{2pt}
Options. A. \figkeyword{Door} B. \figkeyword{Two chairs} C. \figkeyword{Chair and white board}\par\vspace{2pt}
\correctanswer{\georeaderfig{}: A}\hspace{14pt}\correctanswer{3DThinker: A}
\end{minipage}
\par
\includegraphics[width=\linewidth]{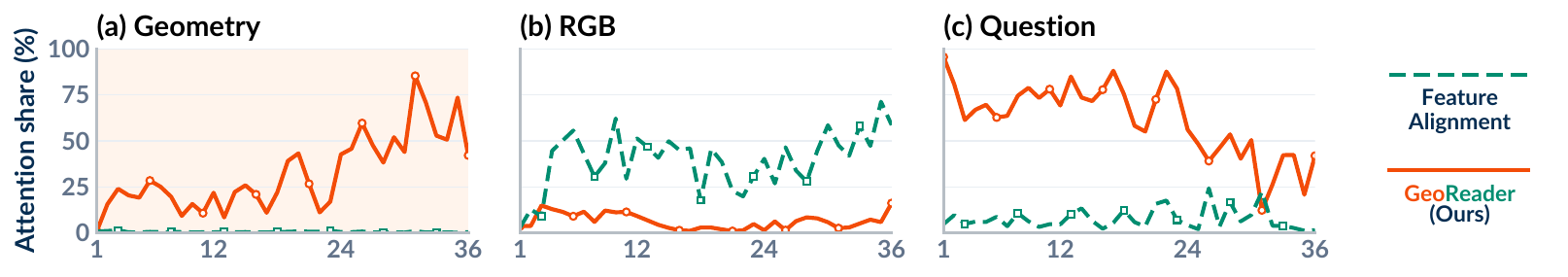}
\end{minipage}};
\end{tikzpicture}\par\vspace{3pt}
\par
\makeatletter\def\@captype{figure}\makeatother
\caption{Answer-token attention across layers on three MindCube-Tiny~\citep{yin2025mindcube} categories. Both models answer every question correctly; \cref{sec:appendix-answer-attention} defines the statistics. In all three cases, answer tokens attend more to geometry in \georeader{} than to latents under feature alignment.}
\label{fig:attention-cases}
\end{minipage}
\end{figure}

We show one example from each MindCube-Tiny~\citep{yin2025mindcube} category: Among, Around, and
Rotation (\cref{fig:attention-cases}).
For each category, we use the first image-matched entry in our
evaluation list, selected before extracting attention and without filtering
by correctness or attention magnitude. The Among example, the black-sneaker question, is also used in
\cref{fig:token-attention}.

We use the pair-supervised Camera-only \georeader{} checkpoint and the
released \texttt{jankin123/3DThinker-Mindcube} checkpoint of
3DThinker~\citep{chen2026think3d} (training step 74,000;
source revision \texttt{c9469e01b719}), the model released after its first
training stage. The latter serves as our representative of feature alignment. Both backbones have 36 language-model layers,
16 query heads, and 2,048-dimensional hidden states.
\georeader{} first reads the RGB and question prefix, then injects one geometric token per input view after the question before answer generation.
3DThinker processes RGB and question tokens followed by
generated latents, intermediate text, and the answer.
We retain each model's native preprocessing and generation procedure.

During each model's greedy generation, we record eager causal
attention at the positions predicting the final answer body, excluding answer tags, and average over query
heads and these positions within each layer. Keys are
grouped into Geometry, RGB, Question, CoT, and Other. Other contains
instruction and chat templates, boundaries, and preceding answer tokens.
Within each layer, attention shares are normalized over Geometry, RGB,
Question, and CoT, excluding Other. CoT remains in the denominator for 3DThinker, although its panel is omitted; \georeader{} has
no CoT body. All 36 layers are retained without smoothing.
\georeader{}'s geometry mass refers to complete geometric tokens, including
their slot embeddings and geometric residuals.

Both models answer all three examples correctly. Averaged over layers, the
geometry share of \georeader{}'s answer positions is $33.8\%$, $27.7\%$ and
$32.3\%$ for Among, Around and Rotation (\cref{fig:attention-cases}), versus $0.2\%$, $0.4\%$ and $0.4\%$ for 3DThinker's latents.

\subsubsection{Additional Qualitative Results}
\label{sec:appendix-predicted-examples}
\paragraph{\cref{fig:qualitative-results}.}
The two MMSI-Bench~\citep{yang2025mmsi} examples use the Camera+Depth
checkpoint at epoch 1, step 6,335; questions and predictions are from saved
evaluation records. Manually marked visible object regions are backprojected
with predicted depth and intrinsics, omitting surface connections across depth
discontinuities. Orange outlines mark these regions; the bathroom
arrow shows the facing direction the question imagines. IMG, Ques.\ and GEO
are image, question and per-view geometry groups. GEO blocks are schematic;
the checkpoint uses one pose and one depth token per view.

\paragraph{\cref{fig:predicted-success,fig:predicted-failure}.}
Panels (a) and (c) of both figures come from MindCube-Tiny~\citep{yin2025mindcube}
and panel (b) from CV-Bench~\citep{tong2024cambrian}; all use DA3 geometry
(\cref{sec:appendix-geometry-sources}). In \cref{fig:predicted-success}a, the
question states that view~2 is turned $90^\circ$ right and view~3 $180^\circ$;
the predicted poses turn them by $98^\circ$ and $177^\circ$. View~3 thus faces
away from view~1, and \georeader{} answers with the window seen in view~3. This
contrasts with \cref{fig:predicted-failure}a, where predicted turns of
$50^\circ$ and $124^\circ$ lead to a wrong answer. In
\cref{fig:predicted-success}b, the predicted median depth is 2.4~m in the
table box and 6.7~m in the bookcase box, and \georeader{} correctly chooses the closer
table, the reference answer.

\subsubsection{Failure Cases}
\label{sec:appendix-failure-cases}
\begin{figure}[p]
\centering
\includegraphics[width=\linewidth]{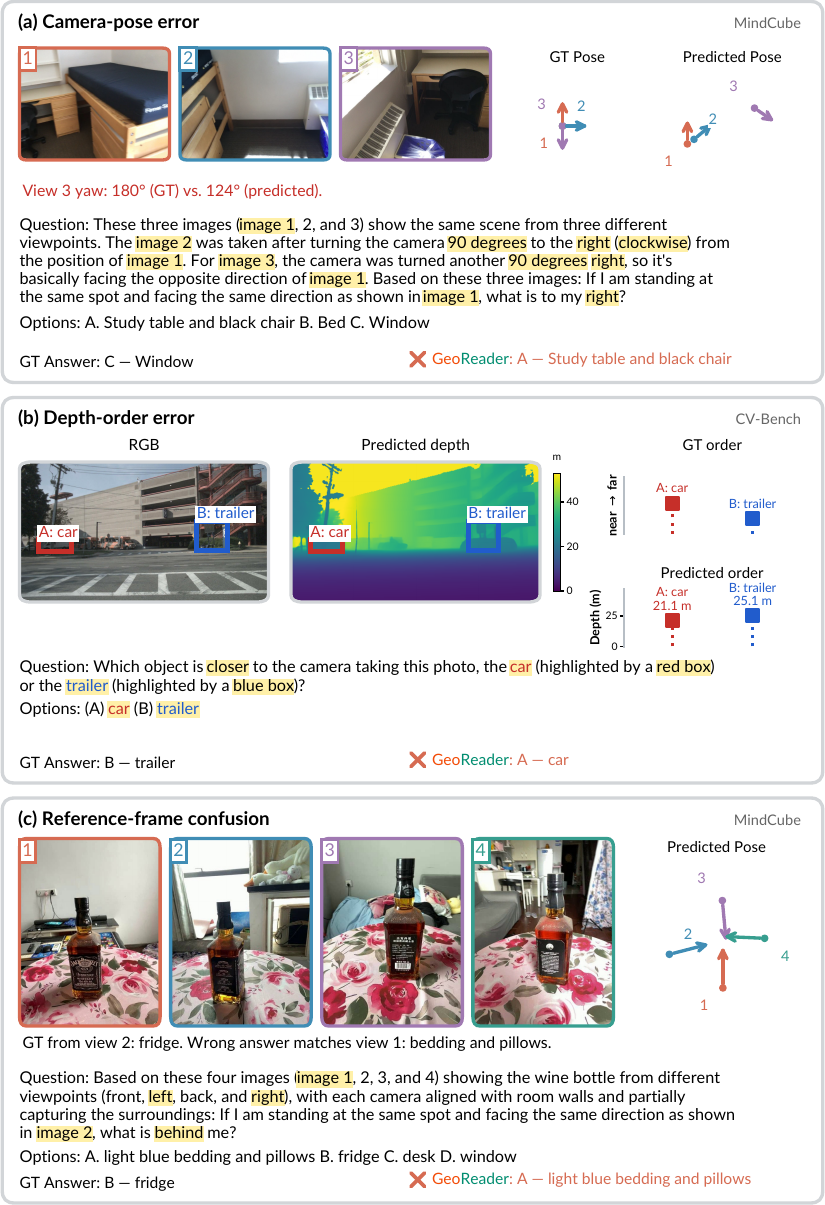}
\caption{Selected failure cases: (a) a wrong response alongside a pose error; (b) reversed predicted depth order for the car/trailer regions; (c) a response matching an alternative reference view. GT pose is annotation-derived; GT order is ordinal. Horizontal depth-plot positions are schematic.}
\label{fig:predicted-failure}
\end{figure}

\cref{fig:predicted-failure} shows three questions that \georeader{} answers incorrectly with predicted geometry, which is wrong in panels (a) and (b) and roughly correct in panel (c).

\paragraph{(1) Camera-pose estimation errors.}
Panel (a) asks what lies to the right when facing the direction of view~1.
The question states that view~2 is turned $90^\circ$ right and view~3
$180^\circ$, but the predicted poses turn them by only $50^\circ$ and
$124^\circ$. View~3 thus becomes the view closest to the right of view~1,
and \georeader{} names the study table and black chair seen in view~3
instead of the window seen in view~2. The two pose plots show the annotated
and predicted view directions.

\paragraph{(2) Depth estimation errors.}
Panel (b) asks whether the car or the trailer is closer to the camera; the
benchmark answer is the trailer. Over the central $40\%$ of each box, the
predicted median depth is $21.1$\,m (car) and $25.1$\,m (trailer). \georeader{} follows this reversed order and chooses the car.

\begin{figure}[p]
\centering
\includegraphics[width=\linewidth]{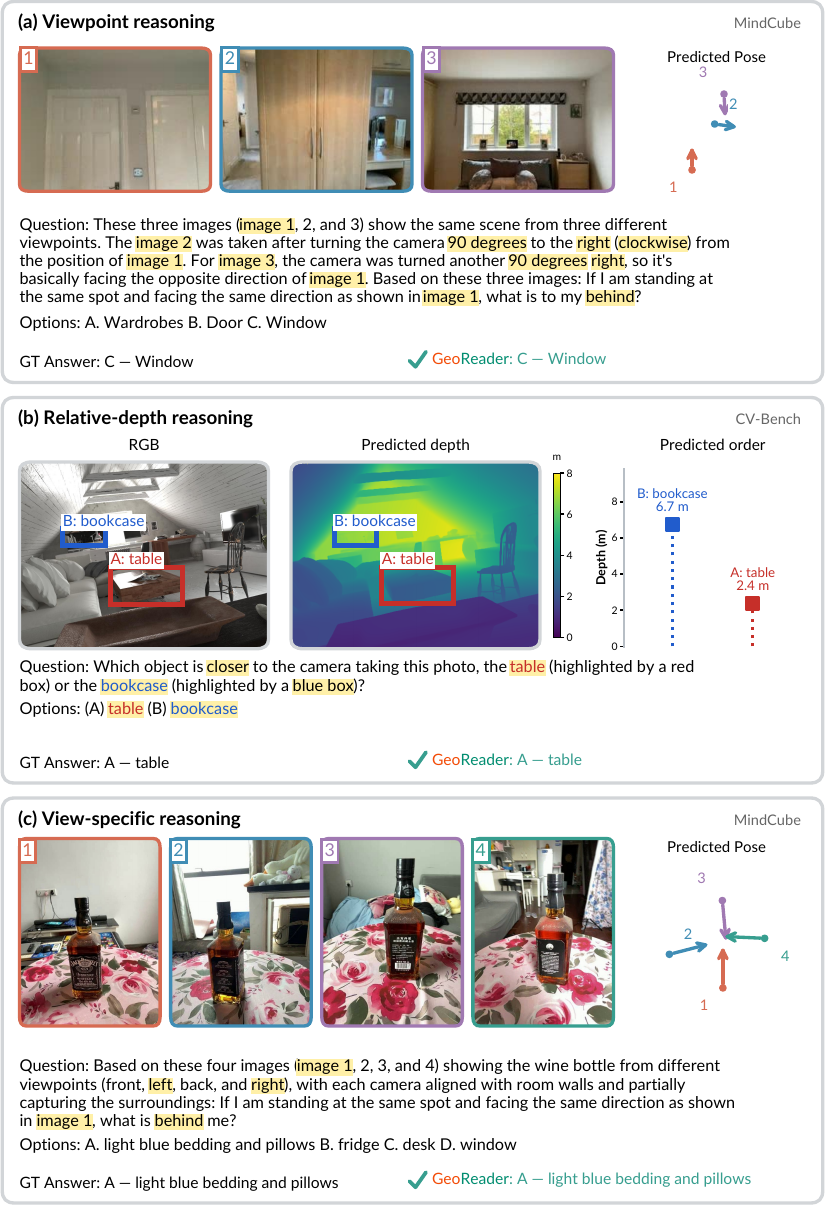}
\caption{Selected correct responses with Depth Anything 3~\citep{lin2025da3} geometry. Camera arrows show poses; depth plots show box-region medians at schematic horizontal positions. Each case includes the question, options, reference answer, and \georeader{} response (\cref{sec:appendix-predicted-examples}).}
\label{fig:predicted-success}
\end{figure}

\paragraph{(3) Reference-frame confusion.}
Panel (c) asks what is behind the viewer at view~2; the reference answer is
the fridge. The predicted pose turns view~2 by $76^\circ$ from view~1, close
to the quarter turn between adjacent views in this setup. \georeader{}
nevertheless answers ``light blue bedding and pillows,'' the reference answer
for the same question at view~1 (cf.\ \cref{fig:predicted-success}c).

\FloatBarrier
\subsection{Public Resources}
\label{sec:appendix-public-resources}
We acknowledge the public data, models and code used in this work.
The dataset list covers all $14$ RGB sources in our approximately $135$K-example
training collection, together with MindCube. QA construction
and the remaining evaluation benchmarks are described in
\cref{sec:appendix-training-mixture,sec:appendix-benchmark-metrics}.
\Cref{sec:appendix-public-data} lists the datasets and
\cref{sec:appendix-public-code} the implementations and models, with their
licenses.

\begingroup
\newcommand{\resitem}[3]{\item #1\footnote{\url{#2}}\nobreak\hspace{0.5em}\dotfill\nobreak\hspace{0.5em}\mbox{#3}}
\subsubsection{Public Datasets}
\label{sec:appendix-public-data}
\begin{itemize}
\setlength{\itemsep}{2pt}\setlength{\parskip}{0pt}
\resitem{ScanNet~\citep{dai2017scannet}}{https://github.com/ScanNet/ScanNet}{ScanNet Terms of Use}
\resitem{ScanNet++~\citep{yeshwanth2023scannetpp}}{https://scannetpp.mlsg.cit.tum.de/scannetpp/}{ScanNet++ Terms of Use}
\resitem{COCO images~\citep{lin2014coco}}{https://cocodataset.org/\#termsofuse}{Image-specific licenses}
\resitem{ARKitScenes~\citep{baruch2021arkitscenes}}{https://github.com/apple/ARKitScenes}{Apple dataset license}
\resitem{Matterport3D~\citep{chang2017matterport3d}}{https://github.com/niessner/Matterport}{Matterport3D Terms of Use}
\resitem{nuScenes~\citep{caesar2020nuscenes}}{https://www.nuscenes.org/terms-of-use}{nuScenes Terms of Use}
\resitem{Spatial-MM~\citep{shiri2024spatialmm}}{https://huggingface.co/datasets/advaitgupta/spatial-mm-obj}{Not specified}
\resitem{AgiBot World~\citep{agibot2025world}}{https://github.com/OpenDriveLab/AgiBot-World}{CC BY-NC-SA 4.0}
\resitem{nuImages~\citep{caesar2020nuscenes}}{https://www.nuscenes.org/terms-of-use}{nuImages Terms of Use}
\resitem{EgoDex~\citep{hoque2026egodex}}{https://github.com/apple-aiml-research/ml-egodex}{CC BY-NC-ND}
\resitem{EPIC-Contact~\citep{bansal2026epiccontact}}{https://huggingface.co/datasets/Sid2697/epic-contact}{CC BY-NC 4.0}
\resitem{DTU MVS~\citep{aanaes2016dtu}}{https://roboimagedata.compute.dtu.dk/?page_id=36}{Not specified}
\resitem{DAVIS images~\citep{ponttuset2017davis}}{https://davischallenge.org/}{Source-video terms}
\resitem{SUN RGB-D~\citep{song2015sunrgbd}}{https://rgbd.cs.princeton.edu/}{Constituent-dataset terms}
\resitem{MindCube~\citep{yin2025mindcube}}{https://huggingface.co/datasets/MLL-Lab/MindCube}{MIT License}
\end{itemize}

\subsubsection{Public Implementations and Models}
\label{sec:appendix-public-code}
Our code builds on 3DThinker~\citep{chen2026think3d}. VLM-3R~\citep{fan2025vlm3r} and SpaceMind~\citep{zhao2025spacemind} inform
the adapted fusion designs in \cref{sec:appendix-adapted-fusion}.
Code and checkpoint licenses are listed separately where they differ.
\begin{itemize}
\setlength{\itemsep}{2pt}\setlength{\parskip}{0pt}
\resitem{3DThinker~\citep{chen2026think3d}}{https://github.com/zhangquanchen/3DThinker}{Apache License 2.0}
\resitem{VLM-3R~\citep{fan2025vlm3r}}{https://github.com/VITA-Group/VLM-3R}{Apache License 2.0}
\resitem{SpaceMind~\citep{zhao2025spacemind}}{https://github.com/RealMikeDuke/SpaceMind}{Apache License 2.0}
\resitem{MindCube evaluation code~\citep{yin2025mindcube}}{https://github.com/mll-lab-nu/MindCube}{MIT License}
\resitem{Depth Anything 3~\citep{lin2025da3}}{https://github.com/ByteDance-Seed/Depth-Anything-3}{Apache License 2.0}
\resitem{Depth Anything 3 Nested~\citep{lin2025da3}}{https://huggingface.co/depth-anything/DA3NESTED-GIANT-LARGE-1.1}{CC BY-NC 4.0}
\resitem{VGGT~\citep{wang2025vggt}}{https://github.com/facebookresearch/vggt}{VGGT License}
\resitem{Qwen2.5-VL-3B-Instruct~\citep{bai2025qwen}}{https://huggingface.co/Qwen/Qwen2.5-VL-3B-Instruct}{Qwen Research License}
\resitem{Qwen2.5-VL-7B-Instruct~\citep{bai2025qwen}}{https://huggingface.co/Qwen/Qwen2.5-VL-7B-Instruct}{Apache License 2.0}
\resitem{Qwen3-VL-2B-Instruct~\citep{bai2025qwen3vl}}{https://huggingface.co/Qwen/Qwen3-VL-2B-Instruct}{Apache License 2.0}
\resitem{Qwen3-VL-4B-Instruct~\citep{bai2025qwen3vl}}{https://huggingface.co/Qwen/Qwen3-VL-4B-Instruct}{Apache License 2.0}
\resitem{Qwen3-VL-8B-Instruct~\citep{bai2025qwen3vl}}{https://huggingface.co/Qwen/Qwen3-VL-8B-Instruct}{Apache License 2.0}
\resitem{Qwen3.5-0.8B~\citep{qwen2026qwen35}}{https://huggingface.co/Qwen/Qwen3.5-0.8B}{Apache License 2.0}
\resitem{Qwen3.5-2B~\citep{qwen2026qwen35}}{https://huggingface.co/Qwen/Qwen3.5-2B}{Apache License 2.0}
\resitem{Qwen3.5-4B~\citep{qwen2026qwen35}}{https://huggingface.co/Qwen/Qwen3.5-4B}{Apache License 2.0}
\resitem{Qwen3.5-9B~\citep{qwen2026qwen35}}{https://huggingface.co/Qwen/Qwen3.5-9B}{Apache License 2.0}
\end{itemize}
\endgroup

\end{document}